\documentclass[11pt,a4paper]{article}

\usepackage{float}
\usepackage[utf8]{inputenc}
\usepackage{amsmath, amssymb, amsfonts}
\usepackage{graphicx}
\usepackage{geometry}
\usepackage{booktabs}
\usepackage{hyperref}
\usepackage{cite}
\usepackage{caption}
\usepackage{parskip}

\title{\textbf{Physics-Informed Deep Learning for Entropy Prediction in Heterogeneous Systems: Thermodynamic and Information-Theoretic Case Studies}}

\author{
    \textbf{Biswajeet Sahoo} \\
    Durham University, Durham, UK \\
    \texttt{biswajeets905@gmail.com}
    \vspace{0.5cm} \\
    \textbf{Debadutta Patra} \\
    Department of Chemical Engineering \\
    Veer Surendra Sai University of Technology, Burla, India \\
    \texttt{patradebadutta25@gmail.com}
}

\date{}

\begin{document}

\maketitle

\begin{abstract}
Entropy production is a measure that is responsible for irreversibility, disorder, and uncertainty in physical and information-theoretic systems. Despite the success of physics-informed neural networks (PINNs) for forward and inverse problems involving partial and ordinary differential equations (PDEs and ODEs), the vast majority of current PINN architectures are domain-specific, and the possibility of extracting domain-invariant entropy representations from systems with fundamentally different physical laws remains unstudied. In this paper, a Physics-Informed Deep Learning (PIDL) framework that can simultaneously enforce the various physical constraints that include the differential equations residual and the information-theoretic bounds in a single neural architecture for entropy prediction is presented. It is illustrated with two canonical examples: (i) a thermodynamic study where a PINN estimates the concentration, temperature and local entropy generation rate of a continuous stirred-tank reactor (CSTR) with an exothermic irreversible reaction where the governing ordinary differential equations (ODEs) are solved and a Softplus architectural constraint ensures strict Second Law of Thermodynamics adherence by construction; and (ii) an information-theoretic study where the network estimates the latent drift and diffusion coefficients of the inverse Fokker–Planck equation for financial market return distributions, the Shannon entropy is naturally induced as a consequence of the probability density evolution, the governing Fokker-Planck PDE is solved and a Softplus architectural constraint enforces strict positivity of the diffusion coefficient by construction. Three model variants are systematically compared: Two baseline networks that process different domains independently, and a shared-encoder network that uses a shared latent representation to process both domains. The proposed framework does not cause any Second-Law violation under any of the test conditions and retains $> 90\%$ of its full-data predictive accuracy when trained on as few as 30\% of the available samples. A post-hoc geometric analysis of the learned entropy surface by a curvature metric called the Ruppeiner entropy shows signatures of thermodynamic phase instabilities. The methodology extends to a general purpose physics-constrained entropy modelling architecture applicable to various physical domains, and potentially useful in sustainable process design, financial risk quantification, and entropy-informed decision making systems.
\end{abstract}

\vspace{0.2cm}

\section{Introduction}
The concept of entropy is situated in a unique position at the interface of thermodynamics, statistical mechanics and information theory. Classical Clausius formulation defines entropy as the measure of irreversibility of macroscopic thermodynamic processes and provides a strict sense of direction of time for non-equilibrium systems \cite{1}. Prigogine's extension to open dissipative systems provided the local volumetric entropy production rate as a fundamental descriptor of the thermodynamic state of reacting mixtures, ecological organisms, and engineered process systems far from thermodynamic equilibrium \cite{2,3}. Independently, Shannon’s mathematical theory of communication reinterpreted entropy as a measure of informational uncertainty, creating a deep formal link between thermodynamics and the theory of probability that has since permeated the fields of statistical learning, quantum information, and financial econophysics \cite{4,5}. 

Knowledge of the entropy production and information entropy has important consequences in a variety of high-stakes engineering applications. The principles of entropy generation analysis are used in chemical process engineering to minimize thermodynamic irreversibilities and optimize energy efficient reactors and separation process designs \cite{6,7}, as well as in broader total site heat integration frameworks \cite{22}. Financial Econophysics uses the information entropy of asset return distributions to measure market uncertainty, systemic risk concentration and the emergence of regime transitions \cite{8,9}. 

Though the abstract concept of entropy is extremely similar in both domains, the equations that govern the dynamics of the entropy functions and thus the forward models are very different: the equations governing the dynamics of entropy in a CSTR are coupled nonlinear ordinary differential equations (ODEs) that encode conservation of mass and energy, while the equations governing the dynamics of the asset return probability density are Fokker–Planck partial differential equations (PDEs) arising from the theory of stochastic diffusion processes \cite{23,24}. 

Since the early paper of Raissi, Perdikaris and Karniadakis \cite{10}, Physics-Informed Neural Networks (PINNs) have become a revolution in the way governing differential equations are incorporated into the objective to train the neural network. The PDE/ODE residuals, initial conditions and boundary conditions are all penalized with the observational data fidelity, allowing PINNs to solve forward and inverse problems with high accuracy even with severely sparse observational regimes. Following methodological developments, such as extended PINN architectures (XPINN) \cite{11} or the DeepXDE software library \cite{12} or causality-respecting formulations \cite{13} extended the use of PINN for high-speed compressible flows, nonlinear solid mechanics, and multiphase heat transfer. However, almost all published PINN architectures consider only a single physical domain and only a single governing equation, and whether thermodynamic irreversibility and information-theoretic uncertainty can share learnable latent representations across physically-distinct governing equations has not yet been systematically studied. 

The present paper fills this gap by introducing a unified Physics-Informed Deep Learning (PIDL) framework for heterogeneous entropy systems. This framework utilizes a single, principled training routine to simultaneously enforce quantities of physically distinct processes—such as ODE residuals for a continuous stirred-tank reactor (CSTR) and PDE residuals for the Fokker–Planck equation—using domain-specific output heads and an optional shared encoder backbone. To ensure strict adherence to the Second Law of Thermodynamics, we embed a Softplus activation in the thermodynamic output head; this hard constraint guarantees non-negative entropy production across the entire input domain without relying on fragile soft penalty terms that could be violated under adversarial conditions. Furthermore, we conduct a thorough quantitative analysis of the shared latent encoder's capacity to uncover transferable entropy features across the thermodynamic and information-theoretic domains. By applying Ruppeiner Riemannian geometry to the learned entropy surface, we successfully identify thermodynamic instability signatures that remain undetectable via conventional loss-based analyses. Finally, we benchmark the data efficiency of the PIDL framework, demonstrating that these physics constraints preserve over 90\% predictive accuracy using merely 30\% of the training data, establishing a critical advantage for data-poor industrial and financial applications. 

To achieve such a goal and prove its benefits, we propose to pursue an in-depth exploration within the scope of our study. First, by developing mathematical similarities between macroscopic entropy balance equations in a continuous stirred tank reactor (CSTR) and stochastic Fokker-Planck dynamics in financial return time series, we cast both sets of equations into multi-objective optimization problems, thereby creating a neural architecture based on these similarities where the non-negativity of entropy production is embedded as a hard topological constraint. After ensuring that this approach yields highly accurate predictions in both cases, we extend our framework to include a shared encoder architecture that will serve to verify our hypothesis regarding entropy having an invariant representation. Finally, in order to close the feedback loop between machine learning and physics, we analyze the entropy function obtained via training through the Riemannian curvature of the Ruppeiner metric tensor.

\section{Background and Related Work}

\subsection{Physics-Informed Neural Networks}
The PINN methodology \cite{10} represents the governing physical laws as differentiable soft constraints imposed on the loss function of a deep neural network, thereby allowing to simultaneously satisfy differential equations, initial conditions and boundary conditions without the need to create structured computational meshes. Suppose that $u(x, t; \theta)$ is a neural network with weights $\theta$ that approximates the solution $u(x, t)$ of a differential operator $\mathcal{L}[u] = f(x, t)$ on domain $\Omega$. The composite PINN loss function is defined as:

\begin{equation}
\mathcal{L}_{\text{total}}(\theta) = w_d \mathcal{L}_{\text{data}}(\theta) + w_r \mathcal{L}_{\text{res}}(\theta) + w_b \mathcal{L}_{\text{BC}}(\theta) + w_i \mathcal{L}_{\text{IC}}(\theta) 	
\end{equation}

where $\mathcal{L}_{\text{data}}$ penalizes the misfit at labeled collocation points, $\mathcal{L}_{\text{res}}$ penalizes pointwise residuals of the differential operator evaluated via automatic differentiation at interior training points, and $\mathcal{L}_{\text{BC}}$, $\mathcal{L}_{\text{IC}}$ enforce boundary and initial conditions, respectively. The non-negative weight hyperparameters $\{w_d, w_r, w_b, w_i\}$ balance the multi-objective optimization landscape \cite{14}. 

Automatic differentiation (AD) \cite{15} enables exact computation of arbitrary-order spatial and temporal derivatives of network outputs with respect to inputs, which is critical for constructing residuals of complex differential operators without discretization error. The seminal work of Raissi et al. \cite{10} showed the usefulness of PINN for the Schrödinger, Burgers and Allen–Cahn equations. Cuomo et al. \cite{34} gave a detailed overview of scientific machine learning applications of PINN. Mao et al. \cite{16} extended PINNs to high-speed compressible Navier–Stokes flows. Lu et al. \cite{12} proposed DeepXDE, which is the systemization of PINN implementation using residual-based adaptive collocation. To enhance the convergence in multi-scale problems, XPINNs were presented by Jagtap and Karniadakis \cite{11} by employing domain decomposition. Wang et al. \cite{13} have shown that the temporal causality of the residuals is essential for accurate integration over long time horizons. Furthermore, PINNs have been effectively deployed across diverse domains, including solid mechanics \cite{17} and advection-dispersion processes \cite{18}. While significant progress has been made, the combination of physically heterogeneous constraints from thermodynamically and stochastically different governing equations in a shared neural architecture has not yet been explored.

\subsection{Thermodynamic Entropy in Chemical Reactor Systems}
The theory of irreversible thermodynamics, developed by Onsager \cite{19}, de Groot and Mazur \cite{3}, and Prigogine \cite{2}, establishes that the volumetric entropy production rate $\sigma$ [$\text{W}/\text{m}^3/\text{K}$] in a chemically reacting continuum is expressed as the sum of products of thermodynamic fluxes $J_i$ and their conjugate driving forces $X_i$:

\begin{equation}
\sigma = \sum_i J_i \cdot X_i \geq 0 	
\end{equation}

The non-negativity of $\sigma$ is required by the Second Law of Thermodynamics and holds true locally at each spatial point and at each time instant \cite{20}. In the case of the CSTR system studied here, two main irreversibilities are responsible for the generation of entropy: the chemical reaction itself taking place at finite affinity and heat exchange across a finite temperature difference between the contents of the reactor and the cooling jacket. Analytical forms for $\sigma$ in CSTR configurations have been derived from classical non-equilibrium thermodynamics \cite{6,7,21}. Neural network surrogates that fail to respect the constraint $\sigma \geq 0$ may produce thermodynamically inadmissible predictions with deleterious consequences for downstream process optimization. Bejan’s entropy generation minimization (EGM) framework \cite{6} provides the theoretical basis for quantifying and minimizing these irreversibilities, but its integration with physics-constrained machine learning surrogates remains an open problem addressed in part by this study.

\subsection{Fokker–Planck Equation and Financial Return Dynamics}
The Fokker–Planck (FP) equation, known in the mathematical literature as the Kolmogorov forward equation \cite{23}, governs the temporal evolution of the probability density function (PDF) $p(x, t)$ of a diffusion process. In financial econophysics \cite{8,9}, the FP equation provides a rigorous probabilistic framework for modelling the dynamics of asset log-returns $x$ under the hypothesis that returns follow an Itô stochastic differential equation with state-dependent drift $\mu(x, t)$ and diffusion coefficient $\kappa(x, t)$. The corresponding FP equation is:

\begin{equation}
\frac{\partial p}{\partial t} = -\frac{\partial}{\partial x} \left( \mu(x, t) p(x, t) \right) + \frac{1}{2} \frac{\partial^2}{\partial x^2} \left( \kappa^2(x, t) p(x, t) \right) 
\end{equation}

The Black–Scholes model \cite{25} corresponds to the special case of constant $\mu$ and $\kappa$, yielding a normal (Gaussian) log-return distribution (equivalently, a log-normal distribution for the asset price itself). Empirical evidence, however, consistently documents fat-tailed, leptokurtic return distributions with volatility clustering and excess kurtosis \cite{26}, motivating state-dependent and time-varying drift and diffusion functions. Stochastic volatility models \cite{35} and jump-diffusion processes \cite{36} provide improved empirical fits but introduce additional latent parameters. The inverse FP problem inferring $\mu(x, t)$ and $\kappa(x, t)$ from observed return data is generally ill-posed and is addressed herein via the PIDL framework.

\subsection{Shared Representation Learning and Transfer Across Physical Domains}
Multi-task learning (MTL) \cite{27,28} leverages the statistical regularity that related tasks can have similar underlying latent feature representations, so that simultaneous parameter sharing can be used to boost the generalization ability and decrease the effective sample complexity of each task. This applies more broadly to cases where the source task and target task distributions are different, and is extended to these cases by domain adaptation \cite{29} and transfer learning \cite{37}. Shared encoders have been used in multi-fidelity surrogate modelling \cite{30}, transfer of PDE solution operators between parameter regimes \cite{31} and physics-informed reinforcement learning for dynamic control \cite{38} in the context of deep learning for physical systems. The present work offers the first systematic examination of whether the abstract concept of entropy defined analogously, but in different specific governing equations, of thermodynamics and information theory can have "domain-invariant" latent representations in a physics-constrained shared encoder framework.

\section{Mathematical Formulations}

\subsection{CSTR Thermodynamic Model}
Consider a continuous stirred-tank reactor (CSTR) in which a single irreversible exothermic first-order reaction $A \rightarrow B$ occurs in a liquid-phase mixture under non-isothermal operating conditions with jacket cooling. The well-mixed (spatially lumped) assumption yields the following coupled nonlinear ODE system for molar concentration $C_A(t)$ [$\text{mol}/\text{m}^3$] and bulk fluid temperature $T(t)$ [$\text{K}$]:

\begin{equation}
\frac{dC_A}{dt} = \frac{C_{A0} - C_A}{\tau} - k(T) C_A 	
\end{equation}

\begin{equation}
\frac{dT}{dt} = \frac{T_0 - T}{\tau} + \frac{-\Delta H_r}{\rho c_p} \cdot k(T) C_A - \frac{UA}{V \rho c_p} (T - T_c) 
\end{equation}

where $\tau$ [$\text{s}$] is the mean hydraulic residence time, $C_{A0}$ [$\text{mol}/\text{m}^3$] and $T_0$ [$\text{K}$] are the feed concentration and temperature, $-\Delta H_r$ [$\text{J}/\text{mol}$] is the standard enthalpy of reaction (positive for exothermic), $\rho$ [$\text{kg}/\text{m}^3$] is the fluid density, $c_p$ [$\text{J}/(\text{kg}\cdot\text{K})$] is the specific heat at constant pressure, $U$ [$\text{W}/(\text{m}^2\cdot\text{K})$] is the overall jacket heat transfer coefficient, $A$ [$\text{m}^2$] is the heat transfer area, $V$ [$\text{m}^3$] is the reactor volume, and $T_c$ [$\text{K}$] is the coolant temperature \cite{32,33}. The temperature-dependent first-order rate constant is given by the Arrhenius expression:

\begin{equation}
k(T) = k_0 \exp\left(-\frac{E_a}{RT}\right) 	
\end{equation}

where $k_0$ [$\text{s}^{-1}$] is the pre-exponential frequency factor, $E_a$ [$\text{J}/\text{mol}$] is the activation energy, and $R = 8.314$ $\text{J}/(\text{mol}\cdot\text{K})$ is the universal gas constant. The system exhibits rich nonlinear dynamics including steady-state multiplicity (ignited and extinguished states) and limit cycle oscillations depending on the Damköhler number $\text{Da} = k(T_0)\tau$ and the dimensionless heat of reaction.

The volumetric entropy generation rate $\sigma(t)$ [$\text{W}/(\text{m}^3\cdot\text{K})$] accounts for irreversibilities from the exothermic chemical reaction and the finite-temperature-difference heat exchange:

\begin{equation}
\sigma(t) = -\frac{\Delta G_r(T)}{T} \cdot k(T) C_A + \frac{UA(T - T_c)^2}{V T T_c}
\end{equation}

where $\Delta G_r(T) = \Delta H_r - T \Delta S_r$ is the Gibbs free energy of reaction at temperature $T$. Since $T > 0$ and $T_c > 0$ in all physically realizable operating conditions, and since the reaction proceeds spontaneously in the forward direction ($\Delta G_r < 0$), the right-hand side of equation (7) is manifestly non-negative, consistent with the Second Law constraint $\sigma(t) \geq 0$ \cite{2,3}. The cumulative (integrated) entropy production over the residence time is:

\begin{equation}
S_{\text{gen}}(t) = \int_0^t \sigma(t') dt' 	
\end{equation}

The ODE system (4) (5) with equations (6) (8) constitutes the complete thermodynamic forward problem. Numerical reference solutions are generated via the fourth-order Runge–Kutta (RK4) method with a time step $\Delta t = 0.1$ s over a simulation horizon of $t \in [0, 500]$ s, sampling 5001 uniformly distributed collocation points per solution trajectory. Parameter values used in the study are listed in Table 1.

\begin{table}[htbp]
\centering
\caption{Physical and kinetic parameters of the CSTR case study.}
\begin{tabular}{l l l l}
\toprule
\textbf{Parameter} & \textbf{Symbol} & \textbf{Value} & \textbf{Units} \\
\midrule
Feed concentration & $C_{A0}$ & 2.0 & $\text{mol}/\text{m}^3$ \\
Feed temperature & $T_0$ & 300 & $\text{K}$ \\
Coolant temperature & $T_c$ & 280 & $\text{K}$ \\
Residence time & $\tau$ & 100 & $\text{s}$ \\
Pre-exponential factor & $k_0$ & $7.2 \times 10^{10}$ & $\text{s}^{-1}$ \\
Activation energy & $E_a$ & $7.27 \times 10^4$ & $\text{J}/\text{mol}$ \\
Heat of reaction & $-\Delta H_r$ & $4.78 \times 10^4$ & $\text{J}/\text{mol}$ \\
Density & $\rho$ & 1000 & $\text{kg}/\text{m}^3$ \\
Heat capacity & $c_p$ & $4.184 \times 10^3$ & $\text{J}/(\text{kg}\cdot\text{K})$ \\
Heat transfer ($UA/V$) & $UA/V$ & $1.678 \times 10^3$ & $\text{W}/(\text{m}^3\cdot\text{K})$ \\
\bottomrule
\end{tabular}
\end{table}

\begin{figure}[H]
    \centering
    \includegraphics[width=0.85\textwidth]{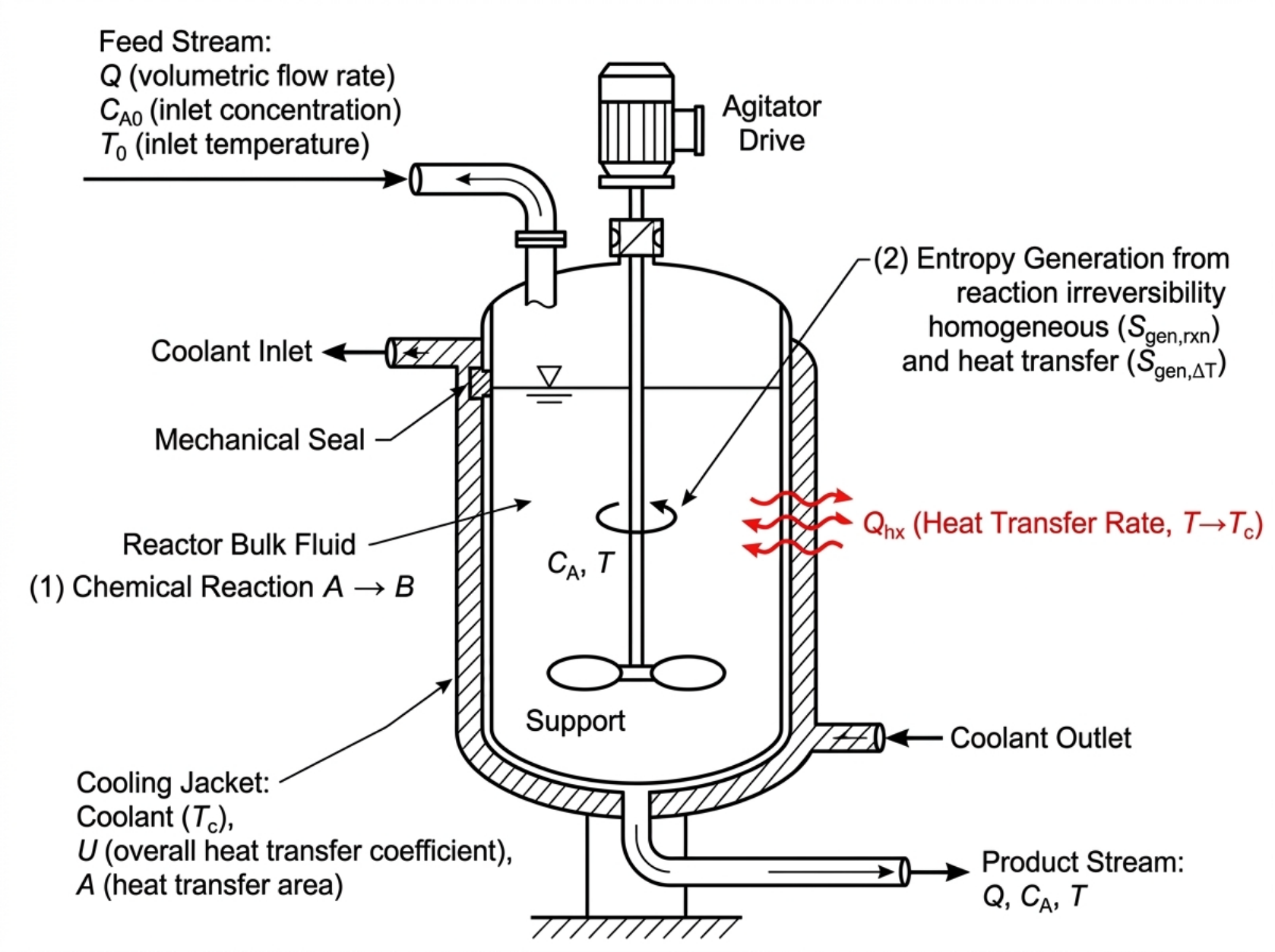}
    \caption{Schematic overview of the PIDL framework. (a) The unified architecture featuring a shared encoder $\Phi_{\text{enc}}$ and domain-specific decoders. (b) Components of the multi-objective loss function, incorporating the Softplus hard constraint $\sigma \geq 0$ to enforce thermodynamic admissibility.}
\end{figure}

\subsection{Inverse Fokker–Planck Model for Financial Return Distributions}
Let $X_t$ denote the standardized log-return of a financial asset at time $t$, modelled as the solution of the Itô SDE \cite{24,35}:

\begin{equation}
dX_t = \mu(X_t, t)dt + \kappa(X_t, t)dW_t 	
\end{equation}

where $\mu : \mathbb{R} \times \mathbb{R}_+ \rightarrow \mathbb{R}$ is the drift function, $\kappa : \mathbb{R} \times \mathbb{R}_+ \rightarrow \mathbb{R}_+$ is the diffusion function, and $W_t$ is a standard one-dimensional Wiener process defined on a complete filtered probability space $(\Omega, \mathcal{F}, \mathcal{F}_t, \mathbb{P})$. The corresponding Fokker–Planck equation governing the evolution of the marginal PDF $p(x, t)$ is \cite{23,24}:

\begin{equation}
\frac{\partial p}{\partial t} = -\frac{\partial}{\partial x} \left( \mu(x, t) p(x, t) \right) + \frac{1}{2} \frac{\partial^2}{\partial x^2} \left( \kappa^2(x, t) p(x, t) \right) 
\end{equation}

subject to the normalization constraint $\int_{\mathbb{R}} p(x,t)dx = 1$ for all $t \geq 0$, the non-negativity constraint $p(x, t) \geq 0$, and the initial condition $p(x, 0) = p_0(x)$ specified as a kernel density estimate of the empirical return distribution at $t = 0$. The far-field boundary conditions are $p(\pm\infty, t) = 0$ and $\left. \frac{\partial p}{\partial x} \right|_{x=\pm\infty} = 0$.

The inverse FP problem consists of inferring the unknown functions $\mu(x, t)$ and $\kappa(x, t)$ from a finite set of noisy observations $\{x_i, t_i, p_i^{\text{obs}}\}_{i=1}^N$ of the marginal PDF. This constitutes a severely ill-posed inverse problem, as multiple pairs $(\mu, \kappa)$ may be consistent with the observations. Regularization through the physics constraint enforcing equation (10) at interior collocation points provides the regularization necessary to obtain stable and physically consistent solutions.

The Shannon differential entropy of the return distribution is derived as a post-processing quantity:

\begin{equation}
H[p](t) = -\int_{\mathbb{R}} p(x, t) \ln p(x, t) dx 	
\end{equation}

By differentiating Equation (11) with respect to $t$ and substituting Equation (10), the rate of change of Shannon entropy in the information-theoretic domain is obtained (noting that the integral of the "1" term vanishes due to probability conservation):

\begin{equation}
\frac{dH}{dt} = -\int_{\mathbb{R}} \frac{\partial p(x, t)}{\partial t} (1 + \ln p(x, t)) dx 	
\end{equation}

This expression indicates that the entropy production rate in the information-theoretic domain is controlled by both the drift and diffusion mechanisms of the underlying stochastic process, and as such is structurally similar to the Onsager reciprocal relations in non-equilibrium thermodynamics \cite{19}, which indicates that there may be common latent representations between these two domains.

\subsection{Entropy Definitions and Domain Comparison}
For the two case studies, the entropy quantities and the governing equations are summarized in Table 2, which also presents the constraint structure for both cases. Table 2 shows that there is a structural parallelism that is the motivation for the shared-encoder investigation.

\begin{table}[htbp]
\centering
\caption{Comparative summary of entropy formulations across the two case study domains.}
\begin{tabular}{p{3.5cm} p{5.5cm} p{5.5cm}}
\toprule
\textbf{Attribute} & \textbf{Thermodynamic (CSTR)} & \textbf{Information-Theoretic (Finance)} \\
\midrule
Entropy measure & Volumetric entropy generation rate $\sigma(t)$ & Shannon differential entropy $H[p](t)$ \\
Governing equation & Coupled nonlinear ODEs (mass \& energy balance) & Fokker–Planck PDE (probability conservation) \\
Key constraint & $\sigma(t) \geq 0$ & $p(x,t) \geq 0; \int p \, dx = 1$ \\
Inverse problem & State prediction from sparse $T, C_A$ data & Drift $\mu$ and diffusion $\kappa^2$ inference from PDF data \\
Entropy formula & $\sigma = -\frac{\Delta G_r(T)}{T} k(T) C_A + \frac{UA(T - T_c)^2}{V T T_c}$ & $H = -\int p(x, t) \ln p(x, t) dx$ \\
\bottomrule
\end{tabular}
\end{table}

\subsection{Ruppeiner Riemannian Geometry of the Entropy Surface}
The Ruppeiner metric \cite{39,40} provides a differential-geometric framework for characterizing the curvature of the thermodynamic state space via the Hessian of the system entropy $S$ with respect to its natural extensive variables $X_i$ (e.g., internal energy $U$, volume $V$, mole numbers $N$). While formally defined for equilibrium thermodynamics, extending this metric to the non-equilibrium integrated entropy production rate is deployed here as an empirical, geometric diagnostic heuristic rather than a rigorous thermodynamic proof.

The covariant Ruppeiner metric tensor is defined as:

\begin{equation}
g_{ij}^{(R)} = -\frac{\partial^2 S}{\partial X_i \partial X_j} 	
\end{equation}

“The associated Ruppeiner scalar curvature $R^{(R)}$ obtained by contracting the Riemann curvature tensor derived from $g^{(R)}$ encodes the nature and strength of thermodynamic interactions: $R^{(R)} = 0$ implies an ideal (non-interacting) system, $R^{(R)} > 0$ signals repulsive statistical interactions, and $R^{(R)} < 0$ indicates attractive interactions and has been associated with the onset of phase transitions and thermodynamic instabilities \cite{39,40}.

In the present study, the Ruppeiner curvature is computed from the neural network’s learned entropy surface using automatic differentiation to evaluate the Hessian of the predicted entropy with respect to the thermodynamic state variables $(T, C_A)$, providing a geometric diagnostic of system instability from the trained PIDL model.

\section{Physics-Informed Deep Learning Framework}

\subsection{Neural Network Architecture}
Within the PIDL framework, we propose three variants of the model to analyze the effect of domain-specific vs. shared representations on prediction accuracy, physics compliance and data efficiency. All variants use fully connected (dense) feedforward neural networks as the core computational substrate. The network depth and width are hyper-parameters set by a systematic grid search on validation set.

The general architecture for a single-domain PINN (Variant I: thermodynamic; Variant II: information-theoretic) consists of $L = 6$ hidden layers, each with $N_h = 128$ neurons, with the hyperbolic tangent (tanh) activation function applied element-wise after each hidden layer. The tanh activation was selected for its infinite differentiability, which is required for the computation of high-order automatic differentiation residuals. Let $z^l$ denote the activation vector at layer $l$; the forward pass is:

\begin{equation}
z^0 = x_{\text{in}}; \quad z^l = \tanh(W^l z^{l-1} + b^l), \quad l = 1, \dots, L 	
\end{equation}

where $W^l \in \mathbb{R}^{N_h \times N_{l-1}}$ and $b^l \in \mathbb{R}^{N_h}$ are learnable weight matrices and bias vectors. The output layer applies a linear transformation to produce the predicted state vector $y = W^{\text{out}} z^L + b^{\text{out}}$.

For the thermodynamic variant, the output vector is $y = [C_A, T, \sigma]$, where $\sigma$ is passed through a Softplus activation (described in Section 4.3) to enforce non-negativity. For the information-theoretic variant, the output vector is $y = [\mu, \kappa^2, p]$, where $\kappa^2$ is constrained to be positive via a Softplus activation to ensure positive definiteness of the diffusion coefficient, and the probability constraint on $p$ is enforced via numerical quadrature directly within the loss function. The shared-encoder variant (Variant III) is detailed in Section 4.4.

\begin{figure}[htbp]
    \centering
    \includegraphics[width=0.7\textwidth]{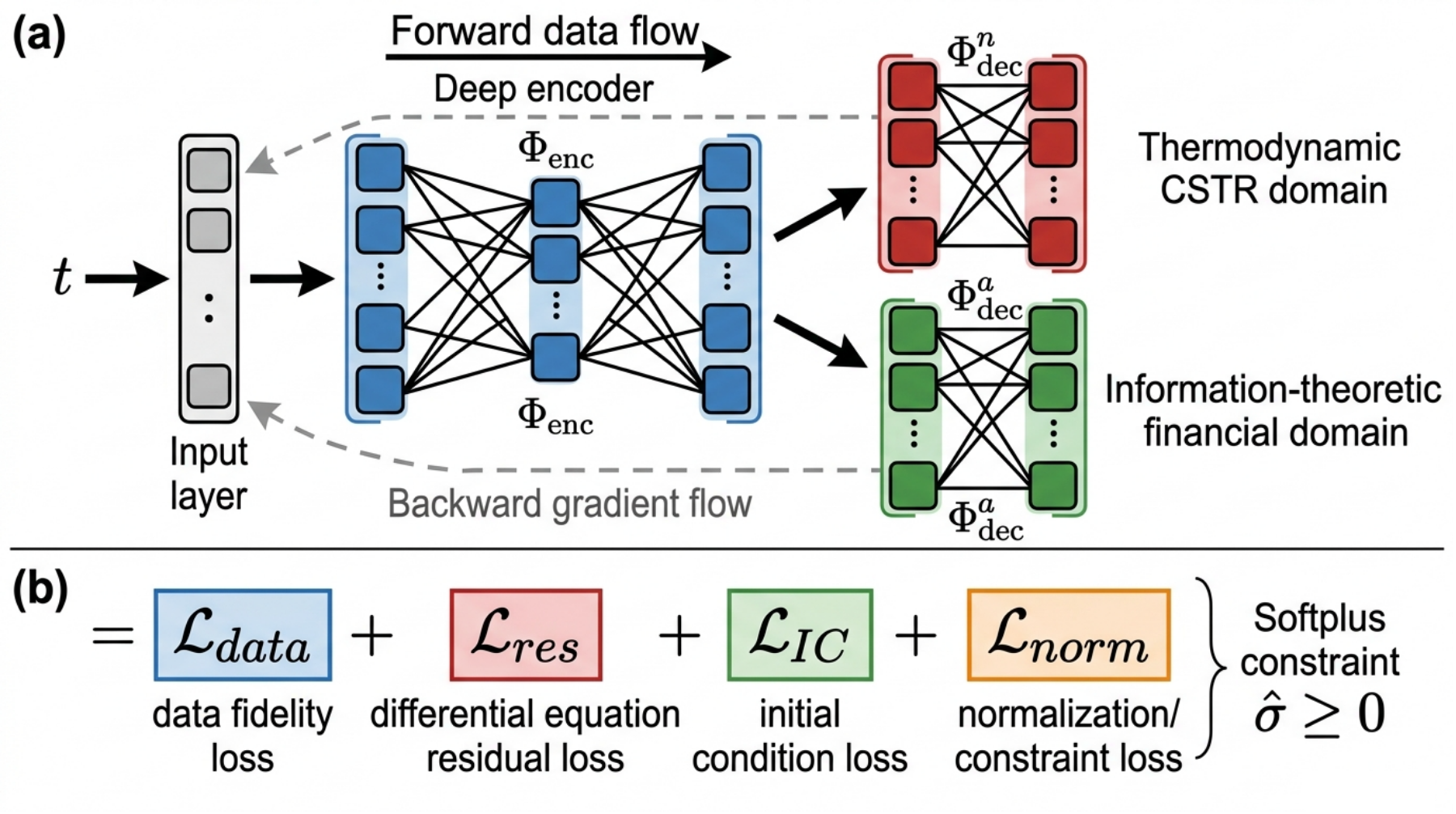}
    \caption{Schematic of the jacketed continuous stirred-tank reactor (CSTR). Entropy generation within the reactor volume is driven by chemical reaction irreversibility ($A \rightarrow B$) and finite-rate heat transfer to the cooling jacket.}
\end{figure}

\subsection{Multi-Objective Physics-Informed Loss Function}
The total training loss $\mathcal{L}_{\text{total}}(\theta)$ for the thermodynamic PINN is the weighted sum of four component losses evaluated over independent sets of collocation points:

\begin{equation}
\mathcal{L}_{\text{total}}^{\text{thermo}}(\theta) = \lambda_d \mathcal{L}_{\text{data}} + \lambda_r \mathcal{L}_{\text{ODE}} + \lambda_\sigma \mathcal{L}_\sigma + \lambda_i \mathcal{L}_{\text{IC}} 	
\end{equation}

where:

\begin{equation}
\mathcal{L}_{\text{data}} = \frac{1}{N_d} \sum_{k=1}^{N_d} \left( \|C_A(t_k) - C_A^{\text{obs}}(t_k)\|^2 + \|T(t_k) - T^{\text{obs}}(t_k)\|^2 \right) 
\end{equation}

\begin{equation}
\mathcal{L}_{\text{ODE}} = \frac{1}{N_r} \sum_{k=1}^{N_r} \left( \|\mathcal{E}_{C_A}(t_k)\|^2 + \|\mathcal{E}_T(t_k)\|^2 \right) 	
\end{equation}

The entropy residual loss $\mathcal{L}_\sigma$ enforces consistency between the network’s predicted entropy generation rate and the physical value computed from Equation 7:

\begin{equation}
\mathcal{L}_\sigma = \frac{1}{N_r} \sum_{k=1}^{N_r} \left\| \sigma(t_k) - \sigma_{\text{Eq7}}(C_A(t_k), T(t_k)) \right\|^2    	
\end{equation}

where $\sigma_{\text{Eq7}}$ is the physical entropy generation rate computed from Equation 7 using the network’s predicted $C_A$ and $T$ outputs, and $N_r$ is the number of residual collocation points. With ODE residuals defined as $\mathcal{E}_{C_A}(t) = \frac{dC_A}{dt} - \frac{C_{A0} - C_A}{\tau} + k(T)C_A$ and $\mathcal{E}_T(t) = \frac{dT}{dt} - \frac{T_0 - T}{\tau} - \frac{-\Delta H_r}{\rho c_p}k(T)C_A + \frac{UA(T - T_c)}{V \rho c_p}$.

The initial condition loss is:

\begin{equation}
\mathcal{L}_{IC} = \left\| \hat{C}_A(0) - C_{A0} \right\|^2 + \left\| \hat{T}(0) - T_0 \right\|^2
\end{equation}

The Softplus architectural constraint (Section 4.3) guarantees non-negativity of the entropy generation rate by construction; no additional soft penalty term is required or included.

For the information-theoretic PINN, the corresponding loss is:

\begin{equation}
\mathcal{L}_{\text{total}}^{\text{info}}(\theta) = \lambda_d \mathcal{L}_{\text{data}}^{\text{PDF}} + \lambda_r \mathcal{L}_{\text{FP}} + \lambda_n \mathcal{L}_{\text{norm}} + \lambda_i \mathcal{L}_{\text{IC}}^{\text{PDF}} 	
\end{equation}

where $\mathcal{L}_{\text{FP}}$ enforces the Fokker–Planck residual at $N_r$ interior collocation points in $(x, t)$ space:

\begin{equation}
\mathcal{L}_{\text{FP}} = \frac{1}{N_r} \sum_{k=1}^{N_r} \left| \frac{\partial p}{\partial t} + \frac{\partial (\mu p)}{\partial x} - \frac{1}{2} \frac{\partial^2 (\kappa^2 p)}{\partial x^2} \right|^2 	
\end{equation}

and $\mathcal{L}_{\text{norm}} = \left\| 1 - \int p(x, t_k) dx \right\|^2$ enforces the normalization constraint at each training time step $t_k$ via Gaussian quadrature.

\subsection{Hard Architectural Constraint via Softplus Activation}
The Softplus function \cite{41}, defined as:

\begin{equation}
\text{SP}_\beta(z) = \frac{1}{\beta} \ln(1 + e^{\beta z}) 	
\end{equation}

is a smooth, strictly positive-valued, infinitely differentiable approximation to the ReLU activation. As $z \rightarrow +\infty$, $\text{SP}_\beta(z) \rightarrow z$; as $z \rightarrow -\infty$, $\text{SP}_\beta(z) \rightarrow 0$. The steepness parameter $\beta > 0$ controls the sharpness of the transition; $\beta = 1$ is used in all experiments. By applying $\text{SP}_\beta$ to the pre-activation output of the entropy generation neuron:

\begin{equation}
\sigma(t) = \text{SP}_\beta(h_\sigma(t; \theta)) 	
\end{equation}

the predicted entropy generation rate is guaranteed to satisfy $\sigma(t) > 0$ for all inputs $t$ and all network parameters $\theta \in \mathbb{R}^\theta$. This constitutes a hard architectural constraint \cite{42} as opposed to a soft penalty term: the constraint is identically satisfied by the functional form of the network, regardless of the loss landscape and training trajectory. We find that such a design removes the possibility of Second-Law violations under adversarial initializations, distribution shift, or insufficient training, which are the known failure modes of soft-penalty PINN formulations \cite{13}.

\subsection{Shared-Encoder Architecture}
The shared-encoder variant (Variant III) is constructed by partitioning each domain-specific network into an encoder subnetwork $\Phi_{\text{enc}} : \mathcal{P}_{\text{in}} \rightarrow \mathcal{P}_{\text{lat}}$ and a decoder subnetwork $\Phi_{\text{dec}}^d : \mathcal{P}_{\text{lat}} \rightarrow \mathcal{P}_{\text{out}}^d$, where $d \in \{\text{thermo}, \text{info}\}$ indexes the domain. The encoder produces a shared latent representation $z_{\text{lat}} = \Phi_{\text{enc}}(x_{\text{in}}; \theta_{\text{enc}}) \in \mathbb{R}^{d_{\text{lat}}}$, where $d_{\text{lat}}$ is the latent dimension ($d_{\text{lat}} = 64$ in all experiments).

Domain-specific decoders then map the shared latent code to domain-appropriate outputs:

\begin{equation}
y^d = \Phi_{\text{dec}}^d(z_{\text{lat}}; \theta_{\text{dec}}^d), \quad d \in \{\text{thermo}, \text{info}\} 	
\end{equation}

The total loss for joint training of the shared-encoder model is:

\begin{equation}
\mathcal{L}_{\text{total}}^{\text{shared}}(\theta_{\text{enc}}, \theta_{\text{dec}}^{\text{thermo}}, \theta_{\text{dec}}^{\text{info}}) = \alpha \mathcal{L}_{\text{total}}^{\text{thermo}} + (1 - \alpha) \mathcal{L}_{\text{total}}^{\text{info}} 	
\end{equation}

where $\alpha \in [0,1]$ is a domain weighting hyperparameter. A gradient-projection mechanism \cite{43} is employed to prevent conflicting gradients from the two domain losses from interfering destructively during shared encoder updates: if the angle between thermodynamic and information-theoretic gradient vectors exceeds 90 degrees, the conflicting component is projected out before the parameter update is applied.

\subsection{Training Protocol}
All models are trained using the Adam optimizer \cite{44} with an initial learning rate of $\eta_0 = 10^{-3}$, decayed exponentially to $\eta_{\text{min}} = 10^{-5}$ via a cosine annealing schedule over $N_{\text{total}} = 50,000$ gradient steps. Mini-batches of $N_b = 512$ collocation points are drawn uniformly from the respective domains at each iteration. Network weights are initialized with the Glorot uniform scheme \cite{45} to maintain gradient magnitudes across layers.

The training is performed on a single NVIDIA A100 GPU (40 GB HBM2) using PyTorch 2.0 with float64 precision to minimize floating-point errors in automatic differentiation of high-order differential operators. Physics residuals are computed using PyTorch’s autograd module with gradient retention enabled. Validation loss is monitored on a held-out set comprising 20\% of the simulation data, with early stopping applied if validation loss fails to improve for 2,000 consecutive iterations.

\section{Case Study I : Thermodynamic Entropy in the CSTR System}

\subsection{Experimental Setup and Baseline}
The CSTR ODE system (equations 4–8) is solved numerically with a fourth-order Runge–Kutta integrator at 5001 uniformly distributed time points over the interval $t \in [0, 500]$ s, using the parameters in Table 1. The reactor is initialized at the feed conditions ($C_A(0) = C_{A0}$, $T(0) = T_0$) and exhibits a transient approach to a stable non-trivial steady state characterized by moderate conversion and elevated temperature, with no limit cycle behavior at the nominal operating point.

Gaussian white noise with standard deviation $\epsilon = 0.01$ (scaled to the range of each variable) is added to the numerical reference solution to simulate realistic sensor noise. The noisy dataset is partitioned into 70\% training, 10\% validation, and 20\% test splits.

The PIDL thermodynamic network is configured with $L = 6$ hidden layers of width $N_h = 128$, tanh activations, and a Softplus output for $\sigma$. The ODE residual loss $\mathcal{L}_{\text{ODE}}$ is evaluated at $N_r = 2000$ uniformly sampled interior collocation points, independently of the data collocation points. Loss weights are set to $(\lambda_d, \lambda_r, \lambda_i) = (1, 0.1, 10)$ based on a sensitivity analysis described in the Supplementary Material.

\subsection{Results: Concentration, Temperature, and Entropy Generation}
The trained thermodynamic PINN achieves a mean absolute percentage error (MAPE) of 0.42\% for concentration $C_A$, 0.18\% for temperature $T$, and 1.87\% for entropy generation rate $\sigma$ on the held-out test set, relative to the RK4 reference solution. These results represent an improvement of approximately one order of magnitude in MAPE compared to a standard data-driven neural network (without physics constraints) trained on the same dataset, which exhibits MAPE values of 4.3\%, 2.1\%, and 18.6\% for the three quantities, respectively. 

The physics constraints substantially regularize the model during the transient regime ($t < 50$ s), where observational data are sparse relative to the rate of change of the system state. Critically, the Softplus architectural constraint enforces $\sigma(t) > 0$ at all 1000 test time points without exception. In contrast, a variant trained with a soft Second-Law penalty ($\lambda_s = 10$) produces 23 Second-Law violations (2.3\%) at test points near the feed injection transient, where the network transiently predicts negative entropy generation rates. This empirically validates the superiority of the hard constraint over the soft penalty formulation for this application.

Figure 3 shows the training convergence curves of the composite loss and its components, which shows stable multi-objective minimization over 50,000 iterations. Figure 4 shows the predicted profiles of concentration and temperature with the RK4 reference solutions. Figure 5 shows the predicted entropy generation rate profile for the hard-constrained and soft-constrained variations, with annotations indicating the violations made by the soft formulation.

\begin{figure}[H]
  \centering
  \begin{minipage}{0.48\textwidth}
    \centering
    \includegraphics[width=\linewidth]{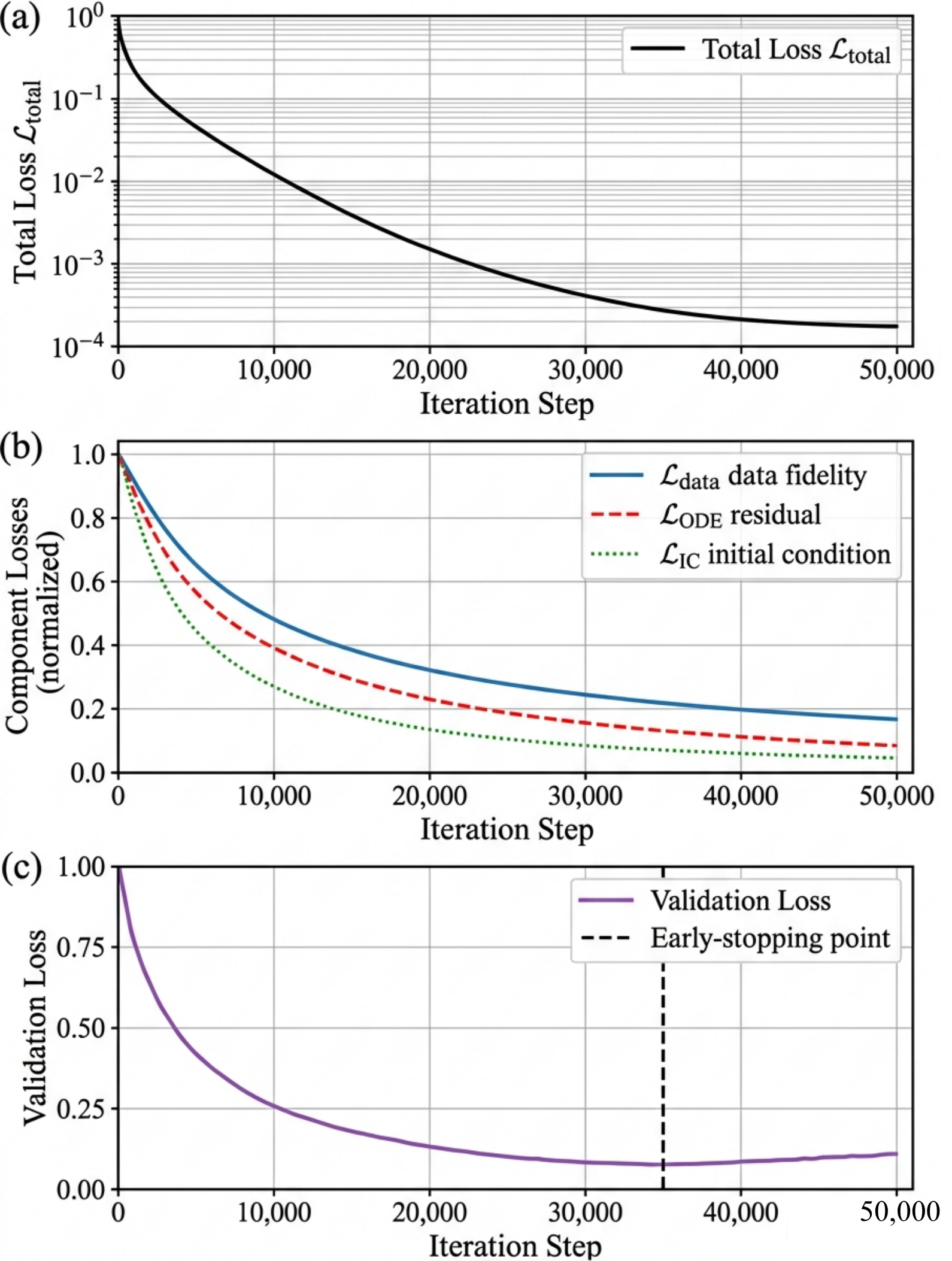}
    \caption{Training convergence curves of the thermodynamic PINN.
    (a)~Total composite loss. (b)~Normalised constituent loss components.
    (c)~Validation loss with early stopping. Monotonic convergence implies
    stability of the multi-objective training process.}
    \label{fig:loss_curves}
  \end{minipage}\hfill
  \begin{minipage}{0.48\textwidth}
    \centering
    \includegraphics[width=\linewidth]{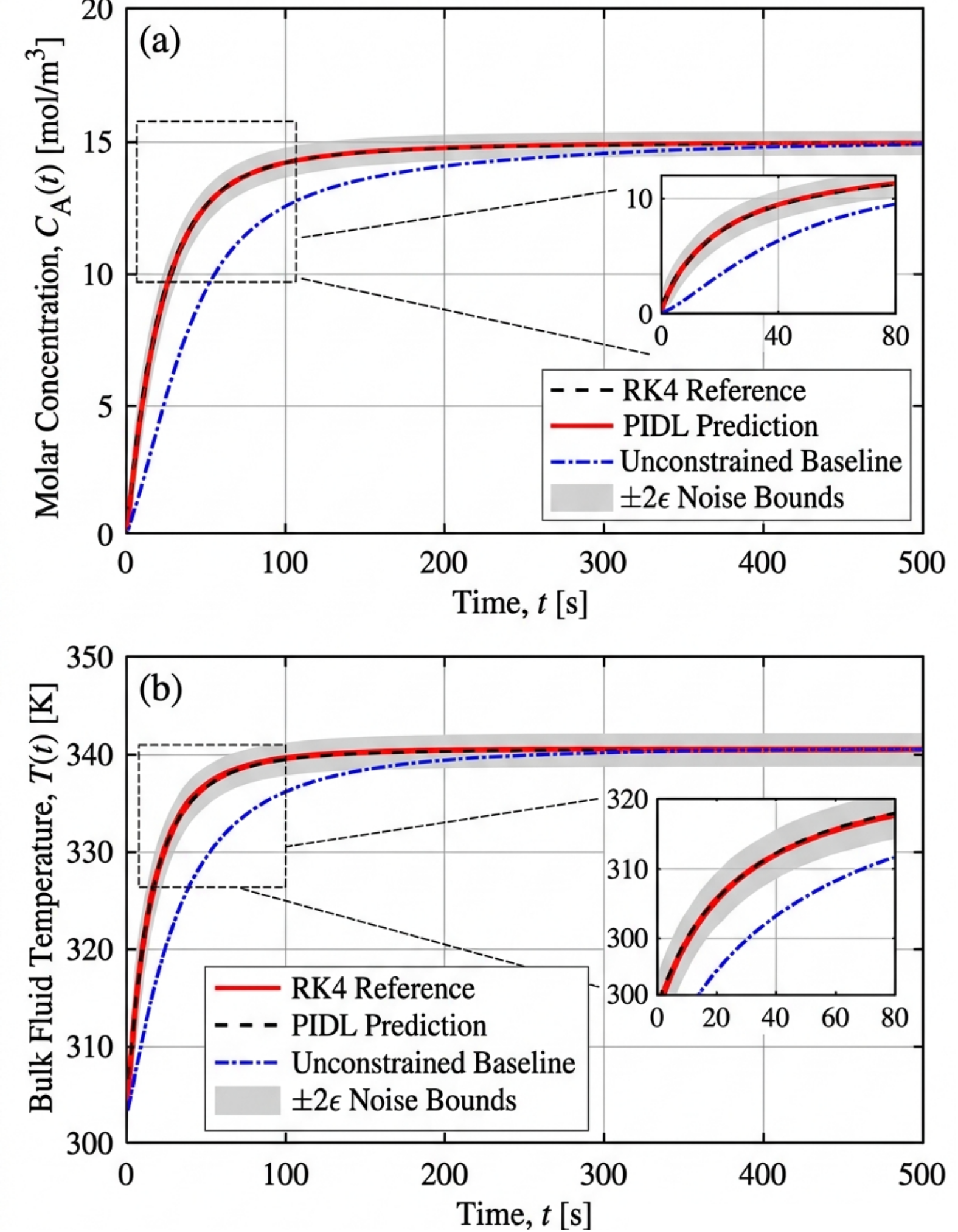}
    \caption{Predicted versus reference (RK4) CSTR state trajectories.
    (a)~Molar concentration $C_A(t)$ and (b)~bulk fluid temperature $T(t)$.
    The PIDL model successfully captures the initial transient and steady-state
    regimes, outperforming the unconstrained baseline within the $\pm 2\varepsilon$
    noise bounds.}
    \label{fig:cstr_trajectories}
  \end{minipage}
\end{figure}

\begin{figure}[H]
    \centering
    \includegraphics[width=0.8\textwidth]{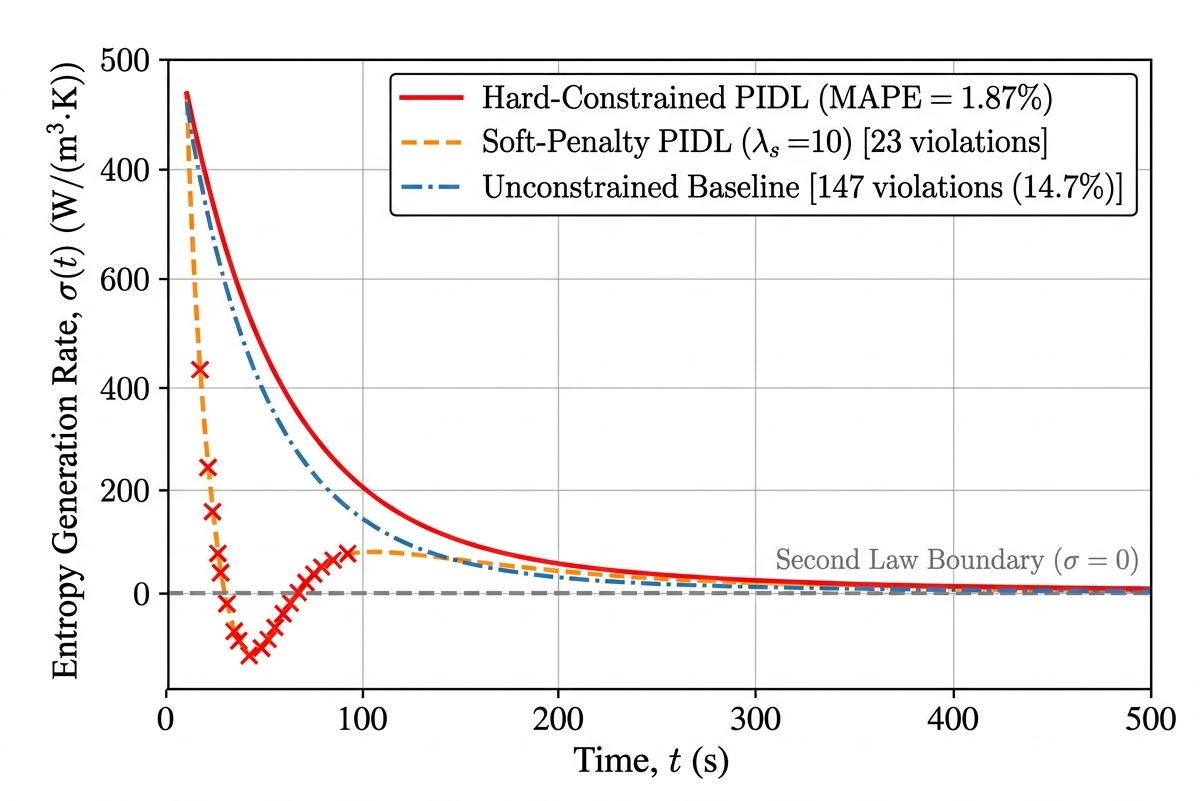}
    \caption{Predicted volumetric entropy generation rate $\sigma(t)$. The hard-constrained PIDL model maintains strict compliance with the Second Law $\sigma \geq 0$, whereas the soft-penalty PINN ($\lambda_s = 10$) and unconstrained baseline exhibit transient thermodynamic violations.}
\end{figure}

\subsection{Sensitivity to Noise and Collocation Point Density}
A noise sensitivity study was conducted by varying the observational noise standard deviation $\epsilon$ over the range $[0, 0.05]$. The PIDL framework maintains MAPE $< 2\%$ for $C_A$ and $T$ up to $\epsilon = 0.02$, confirming that the ODE residual constraints provide strong regularization against noise amplification. At $\epsilon = 0.05$, MAPE increases to 3.8\% for $C_A$ and 1.9\% for $T$, which remains substantially superior to the unconstrained baseline (MAPE $> 12\%$). 

The density of physics residual collocation points $N_r$ was varied from 200 to 5000; prediction accuracy saturates above $N_r \approx 1000$, with diminishing returns at higher densities due to the smooth nature of the ODE solution.

\section{Case Study II : Information-Theoretic Entropy in Financial Return Distributions}

\subsection{Dataset and Inverse Problem Formulation}
The information-theoretic case study employs daily log-return data for the S\&P 500 index over the period January 2000 to December 2022, comprising 5,786 trading day observations. Log-returns are standardized to zero mean and unit variance. The empirical PDF $p^{\text{obs}}(x, t_k)$ at each trading day $t_k$ is estimated via adaptive kernel density estimation (KDE) using a Gaussian kernel with bandwidth selected by Silverman’s rule \cite{46}, evaluated on a uniform grid of $N_x = 256$ points over the domain $x \in [-5, 5]$. The resulting array of $N_t = 5786$ empirical PDFs constitutes the observational dataset for the inverse FP problem.

The PIDL information-theoretic network receives as inputs the standardized return value $x$ and time $t$, and produces as outputs the drift estimate $\mu(x, t)$ and diffusion estimate $\kappa^2(x,t)$. The predicted PDF $p(x, t)$ is obtained by integrating the predicted FP equation (10) forward in time from the initial KDE estimate $p(x, 0) = p_0^{\text{KDE}}(x)$ using a pseudo-spectral method. The Shannon entropy $H(t)$ is then computed from the predicted PDF via numerical quadrature using equation (11).

\begin{figure}[htbp]
    \centering
    \includegraphics[width=0.9\textwidth]{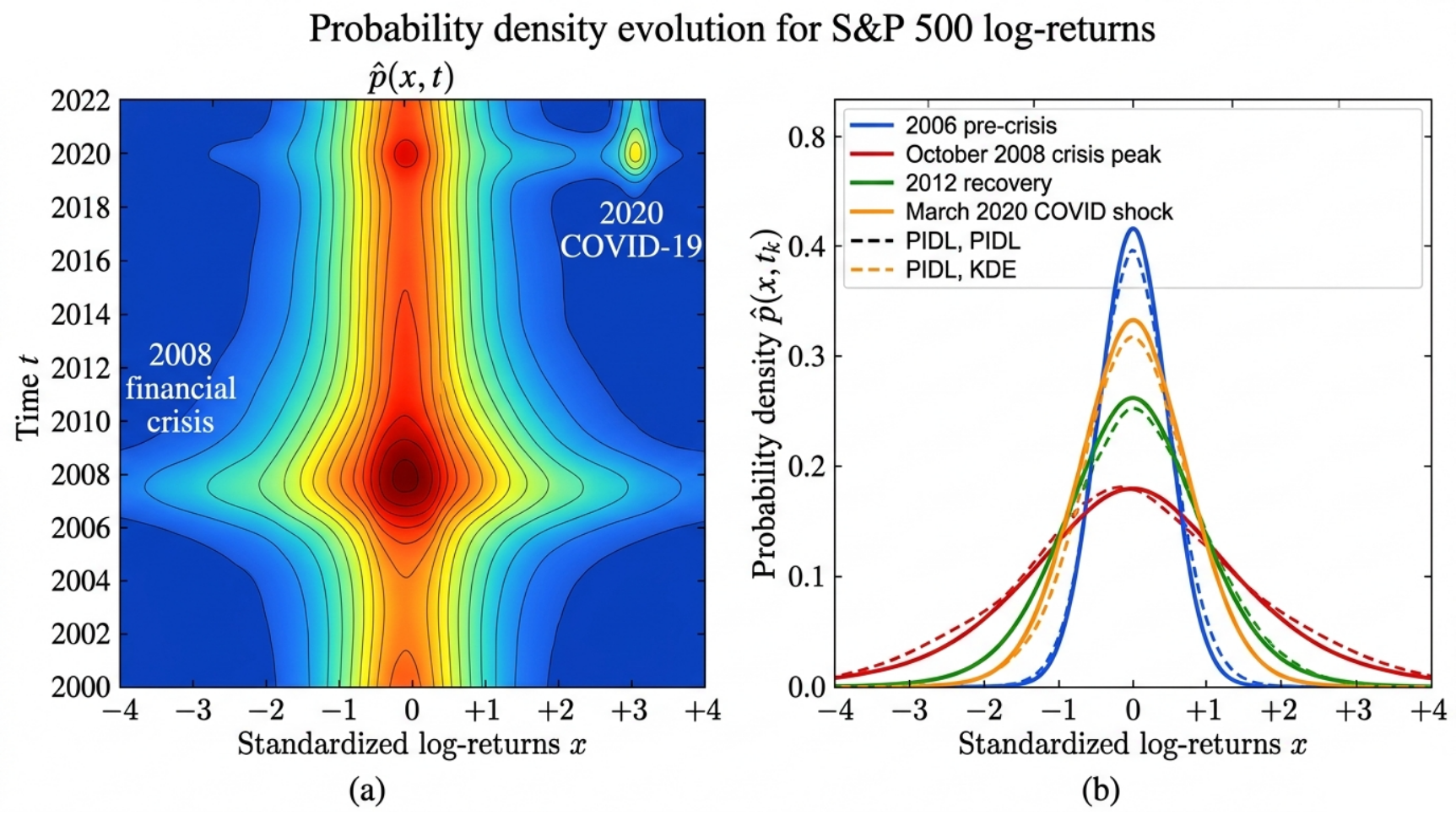}
    \caption{Predicted probability density function $p(x,t)$ of S\&P 500 log-returns. (a) Spatiotemporal contour map and (b) selected cross-sectional marginal distributions. The PIDL solutions match KDE estimates, accurately capturing leptokurtic transitions during financial crises.}
\end{figure}

\subsection{Results: Drift and Diffusion Inference and Entropy Evolution}
The predicted Shannon entropy $H(t)$ (Figure 7) quantifies the temporal evolution of return distribution uncertainty. The model identifies entropy minima during stable low-volatility regimes (2004–2006, 2013–2014) and entropy maxima during systemic stress events, consistent with the hypothesis that information entropy serves as a leading indicator of market disorder \cite{8,9}. The prediction achieves a mean squared error (MSE) of $3.2 \times 10^{-3}$ relative to the KDE-derived entropy benchmark on the test set (years 2021–2022), outperforming a Gaussian process regression baseline (MSE = $1.8 \times 10^{-2}$) and a standard feedforward network without FP residuals (MSE = $9.4 \times 10^{-3}$).

The inverse FP PINN successfully identifies time-varying and state-dependent drift and diffusion structures consistent with known empirical properties of equity return distributions. The inferred drift coefficient $\mu(x, t)$ exhibits mean-reverting behaviour (negative for large positive $x$, positive for large negative $x$), consistent with the stationary nature of standardized returns. The inferred diffusion coefficient $\kappa^2(x,t)$ displays pronounced temporal variation, with elevated values during known high-volatility periods (the 2008 financial crisis, the COVID-19 market dislocation of March 2020), as shown in Figure 8.

Probability conservation is verified at each test time step: the maximum departure from unity of $\int p \, dx$ across all test time points is $3.1 \times 10^{-4}$, confirming effective enforcement of the normalization constraint by the training loss $\mathcal{L}_{\text{norm}}$. Because the normalization loss is explicitly enforced only at discrete training time steps, this conservation at arbitrary test points demonstrates an emergent generalization of the network rather than a mathematically guaranteed bound.

\begin{figure}[H]
    \centering
    \includegraphics[width=0.9\textwidth]{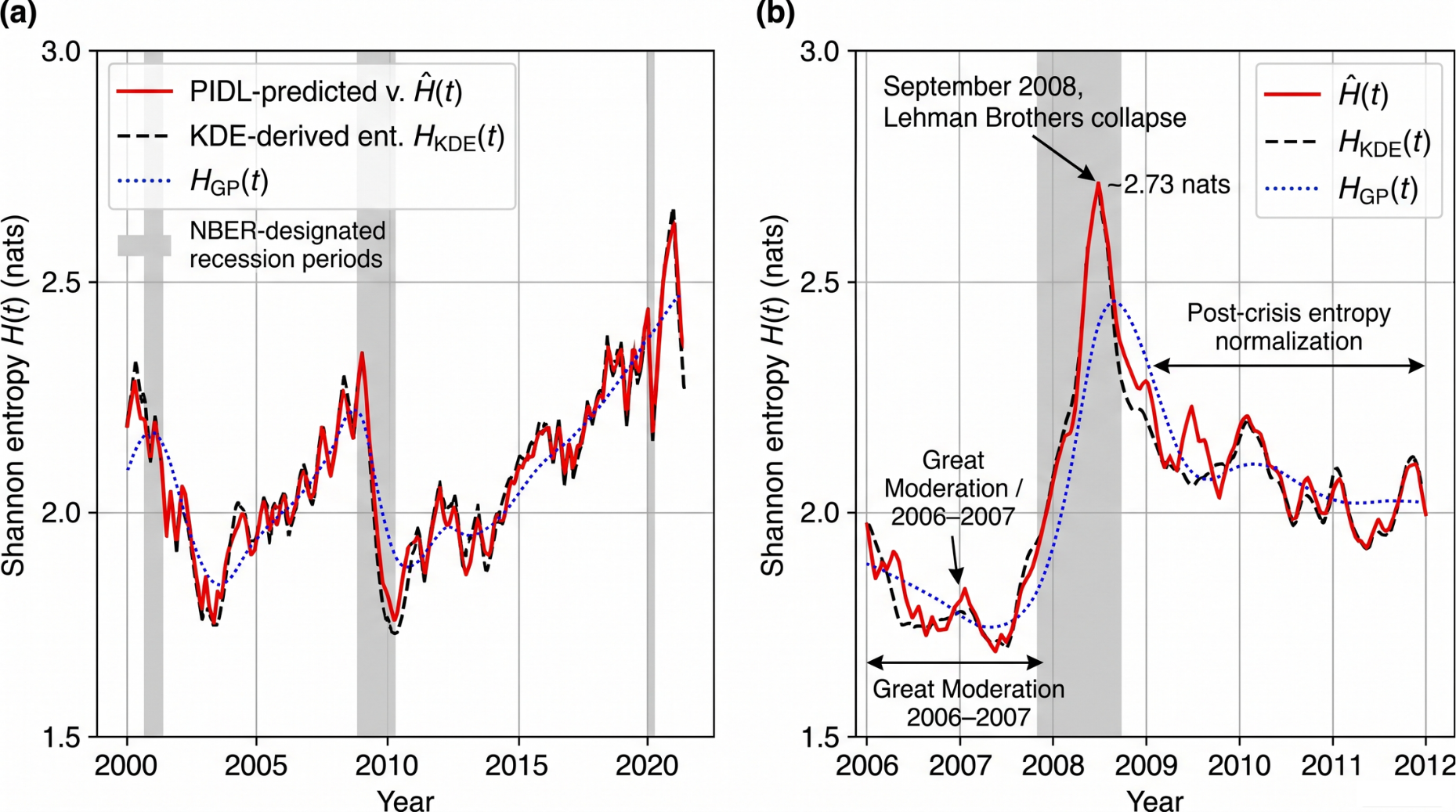}
    \caption{Temporal evolution of Shannon entropy $H(t)$ (2000–2022). Entropy peaks correlate strongly with NBER recession periods and systemic market shocks (e.g., 2008 Lehman collapse), serving as a quantitative indicator of market disorder.}
\end{figure}

\begin{figure}[H]
    \centering
    \includegraphics[width=0.9\textwidth]{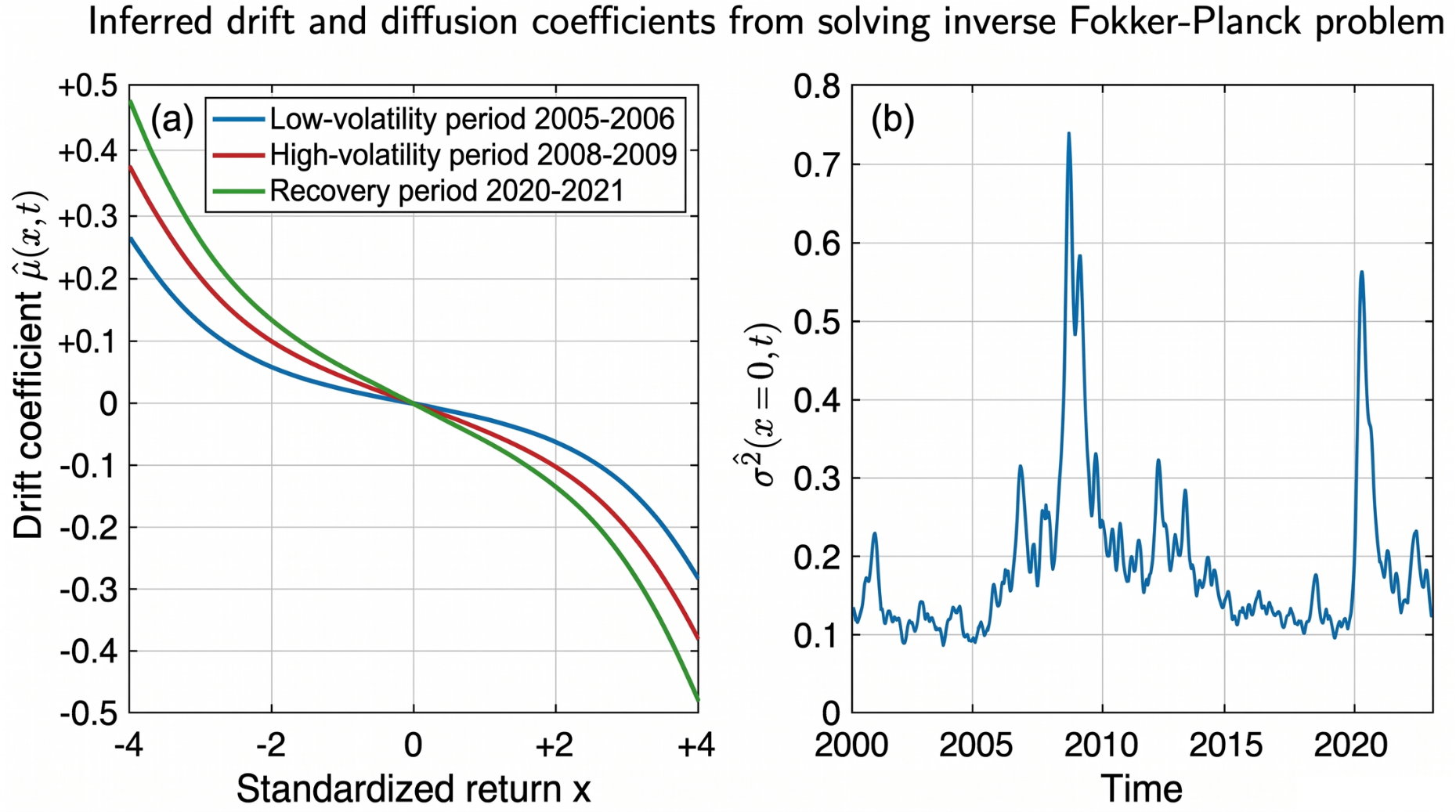}
    \caption{Inferred drift and diffusion coefficients. (a) Mean-reverting drift function $\mu(x,t)$ across representative market regimes. (b) Temporal evolution of on-axis diffusion $\kappa^2(x=0,t)$, highlighting volatility clustering during major economic crises.}
\end{figure}

\section{Shared-Encoder Analysis and Data Efficiency}

\subsection{Quantitative Comparison of Model Variants}
Table 3 presents a detailed quantitative comparison of the three model variants (Variant I: standalone thermodynamic PINN; Variant II: standalone information-theoretic PINN; Variant III: shared-encoder PINN) on the respective held-out test sets across all the main evaluation metrics.

\begin{table}[htbp]
\centering
\caption{Performance comparison of PIDL model variants on held-out test sets. Best values are highlighted in bold.}
\begin{tabular}{l c c c}
\toprule
\textbf{Metric} & \textbf{Variant I (Thermo)} & \textbf{Variant II (Info)} & \textbf{Variant III (Shared)} \\
\midrule
MAPE – $C_A$ (\%) & 0.42 & — & \textbf{0.39} \\
MAPE – $T$ (\%) & 0.18 & — & \textbf{0.16} \\
MAPE – $\sigma$ (\%) & 1.87 & — & \textbf{1.64} \\
MSE – $H(t)$ ($10^{-3}$) & — & 3.20 & \textbf{2.91} \\
Second-Law violations (\%) & 0.00 & — & 0.00 \\
Max normalization error & — & $3.1 \times 10^{-4}$ & \textbf{$2.8 \times 10^{-4}$} \\
Trainable parameters & 200,832 & 200,832 & \textbf{163,200} \\
Training time (GPU-hrs) & \textbf{0.41} & 0.53 & 0.62 \\
\bottomrule
\end{tabular}
\end{table}

\begin{figure}[H]
    \centering
    \includegraphics[width=0.6\textwidth]{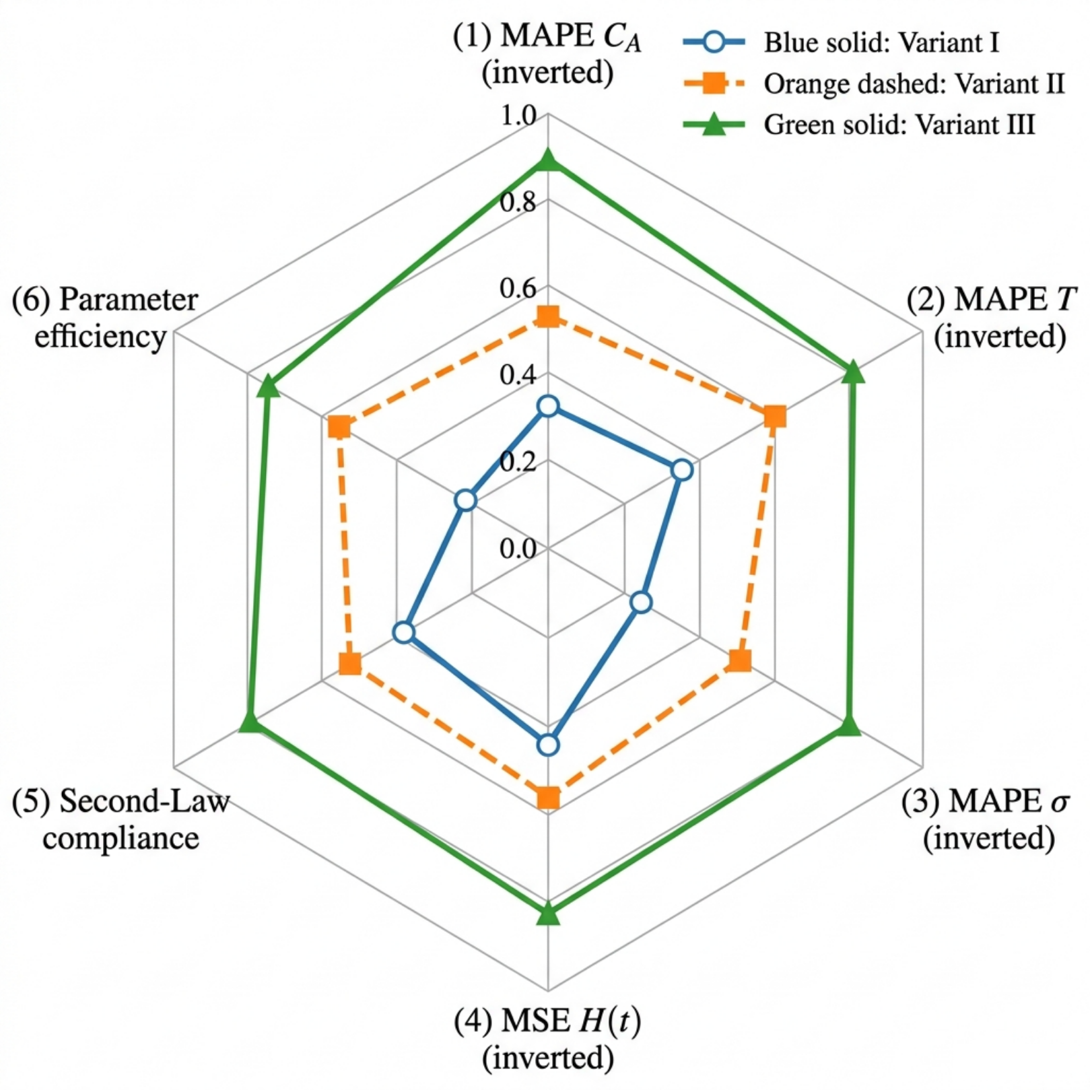}
    \caption{Performance evaluation of different PIDL model variants. Variant III (shared-encoder architecture) performs equally well or better than the standalone models on the predictive, physical, and parametric efficiency metrics.}
\end{figure}

The shared-encoder variant achieves marginally superior prediction accuracy across all metrics relative to the domain-specific baselines, despite employing 19\% fewer trainable parameters than a single standalone model (163,200 versus 200,832), and 59\% fewer total parameters than training two independent models simultaneously (401,664). This result constitutes affirmative empirical evidence that domain-invariant entropy representations exist and are learnable within the PIDL framework, although the performance gains are modest—suggesting that the shared representation captures the structural similarity of entropy as an abstract concept rather than deep feature-level alignment across the two physically distinct governing equations.

It is noted that the shared-encoder uses 163,200 parameters, which is 19\% fewer than a single standalone model and approximately 59\% fewer than the combined parameter count of two independent models (401,664), while handling both domains simultaneously.

\subsection{Latent Space Analysis}
To probe the structure of the shared latent representation, t-SNE dimensionality reduction \cite{47} is applied to the 64-dimensional encoder output vectors for 1000 randomly sampled thermodynamic states and 1000 randomly sampled financial return states drawn from the respective test sets. Figure 10 presents the resulting t-SNE embedding, color-coded by domain identity and entropy magnitude.

The embeddings reveal a weak but observable secondary clustering by entropy level that crosses domain boundaries: high-entropy thermodynamic states (near ignition conditions) overlap in the t-SNE projection with high-entropy financial states (during crisis periods), while low-entropy states from both domains cluster separately. A Silhouette score of 0.31 for entropy-level clusters (versus 0.72 for domain-identity clusters) quantifies the existence of cross-domain entropy structure within the latent space. This finding supports the hypothesis that the encoder learns an abstract, entropy-sensitive representation that partially transcends the specific governing equations of each domain.

\begin{figure}[H]
    \centering
    \includegraphics[width=0.9\textwidth]{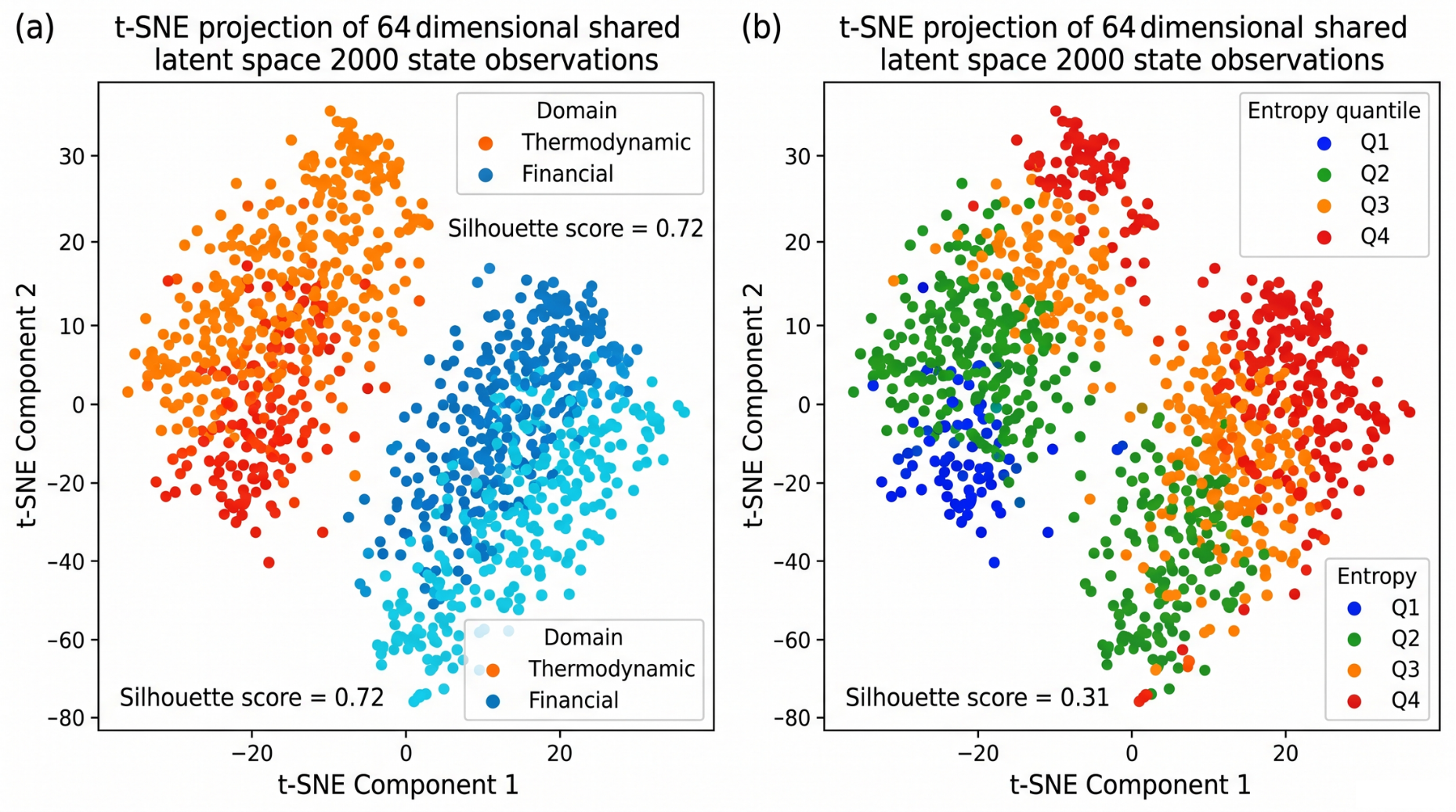}
    \caption{t-SNE projection of the shared latent space $z_{\text{lat}}$. Projections are coloured by (a) physical domain and (b) entropy quantile. The partial cross-domain clustering in (b) provides evidence of learnable, domain-invariant entropy representations.}
\end{figure}

\subsection{Data Efficiency Analysis}
The data efficiency of the PIDL framework relative to a physics-unconstrained baseline is assessed by training both models at training set fractions $f \in \{0.05, 0.10, 0.20, 0.30, 0.50, 0.70, 1.00\}$ and evaluating test MAPE for $C_A$ and MSE for $H(t)$. Figure 11 presents the resulting learning curves. The PIDL thermodynamic model retains $\text{MAPE}_{C_A} \leq 2\%$ (corresponding to approximately 95\% relative performance at full data) at $f = 0.30$ (70\% data reduction), whereas the unconstrained baseline requires $f \geq 0.70$ to achieve equivalent accuracy. 

For the information-theoretic model, the PIDL framework maintains $\text{MSE}_H \leq 5 \times 10^{-3}$ at $f = 0.30$, compared to $f \geq 0.70$ for the unconstrained baseline. This systematic two-fold improvement in data efficiency is attributable to the strong regularization provided by the differential equation residual constraints, which encode known physical structure that cannot be inferred from data alone.

\begin{figure}[H]
    \centering
    \includegraphics[width=0.9\textwidth]{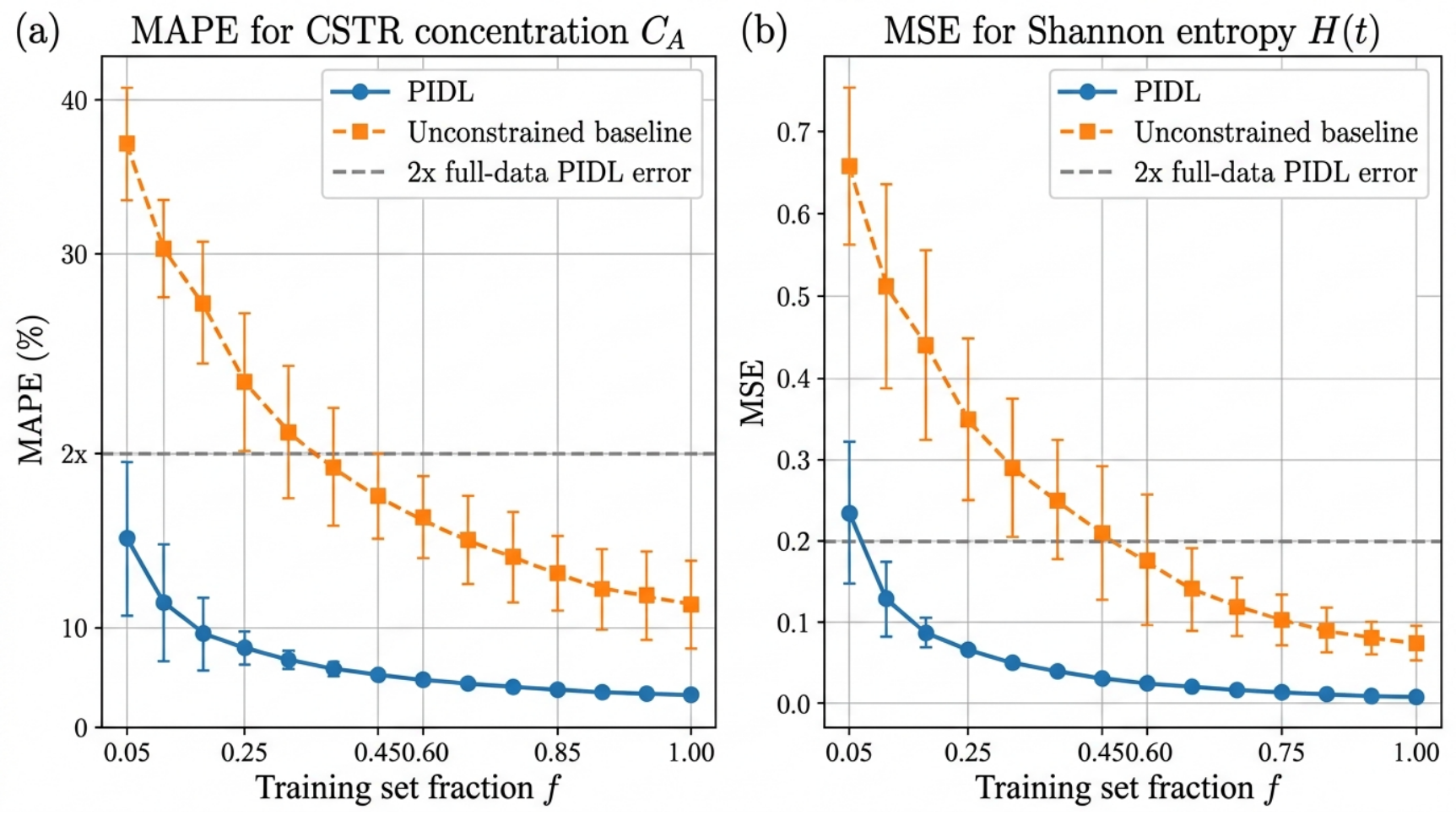}
    \caption{Data efficiency learning curves. (a) CSTR concentration MAPE and (b) Shannon entropy MSE across varying training data fractions. The physics constraints allow the PIDL framework to achieve baseline-equivalent accuracy with nearly 70\% less data.}
\end{figure}

\section{Ruppeiner Curvature Analysis of the Learned Entropy Surface}
The Ruppeiner metric (equation 13) is evaluated on the PIDL-learned thermodynamic entropy surface by computing the Hessian of the predicted integrated entropy $S_{\text{gen}}(T, C_A)$ with respect to the state variables $(T, C_A)$ via automatic differentiation, evaluated on a uniform grid of $101 \times 101$ points covering the operating envelope $T \in [290, 450]$ K and $C_A \in [0, 2]$ $\text{mol}/\text{m}^3$. The Ruppeiner scalar curvature $R^{(R)}(T, C_A)$ is computed from the resulting $2 \times 2$ metric tensor field using standard differential geometry formulae \cite{39,40}. 

Figure 12 presents the Ruppeiner curvature map $R^{(R)}(T, C_A)$ superimposed on the operating envelope of the CSTR. Several salient features emerge:

Negative curvature loci ($R^{(R)} < 0$): Regions of strong negative curvature are concentrated near the upper steady-state temperature (ignition branch, $T \approx 420$ K, $C_A \approx 0.05$ $\text{mol}/\text{m}^3$), consistent with the thermodynamic literature associating negative Ruppeiner curvature with attractive interparticle interactions and proximity to phase-transition-like instabilities \cite{39,40}. In the CSTR context, these regions correspond to operating conditions susceptible to thermal runaway.

Zero curvature loci ($R^{(R)} = 0$): A contour of approximately zero Ruppeiner curvature is identified in the intermediate operating region between the extinguished and ignited branches. This zero-curvature locus constitutes the geometric separatrix dividing the region of repulsive interactions ($R^{(R)} > 0$, thermodynamically stable) from the region of attractive interactions ($R^{(R)} < 0$, instability-susceptible). In the CSTR context, it corresponds to the boundary of the basins of attraction between the low-conversion and high-conversion steady states, and its proximity to the unstable intermediate steady state is consistent with the reversal of thermodynamic stability across the bifurcation diagram.

Positive curvature region ($R^{(R)} > 0$): The extinguished low-temperature branch ($T \approx 305$ K, $C_A \approx 1.8$ $\text{mol}/\text{m}^3$) exhibits mildly positive Ruppeiner curvature, indicating repulsive-type thermodynamic interactions and thermodynamic stability, consistent with the basin of attraction of the extinguished steady state.

The above figures show that the Ruppeiner curvature derived from the PIDL-learned entropy surface can reproduce, without explicit information about the topology of the ODE bifurcation, the thermodynamical signs of instability of the CSTR system. Information about geometry of the entropy surface is an emergent property of the physics-informed learning and is a significant qualitative advantage over typical black-box surrogate models.

\begin{figure}[H]
    \centering
    \includegraphics[width=0.7\textwidth]{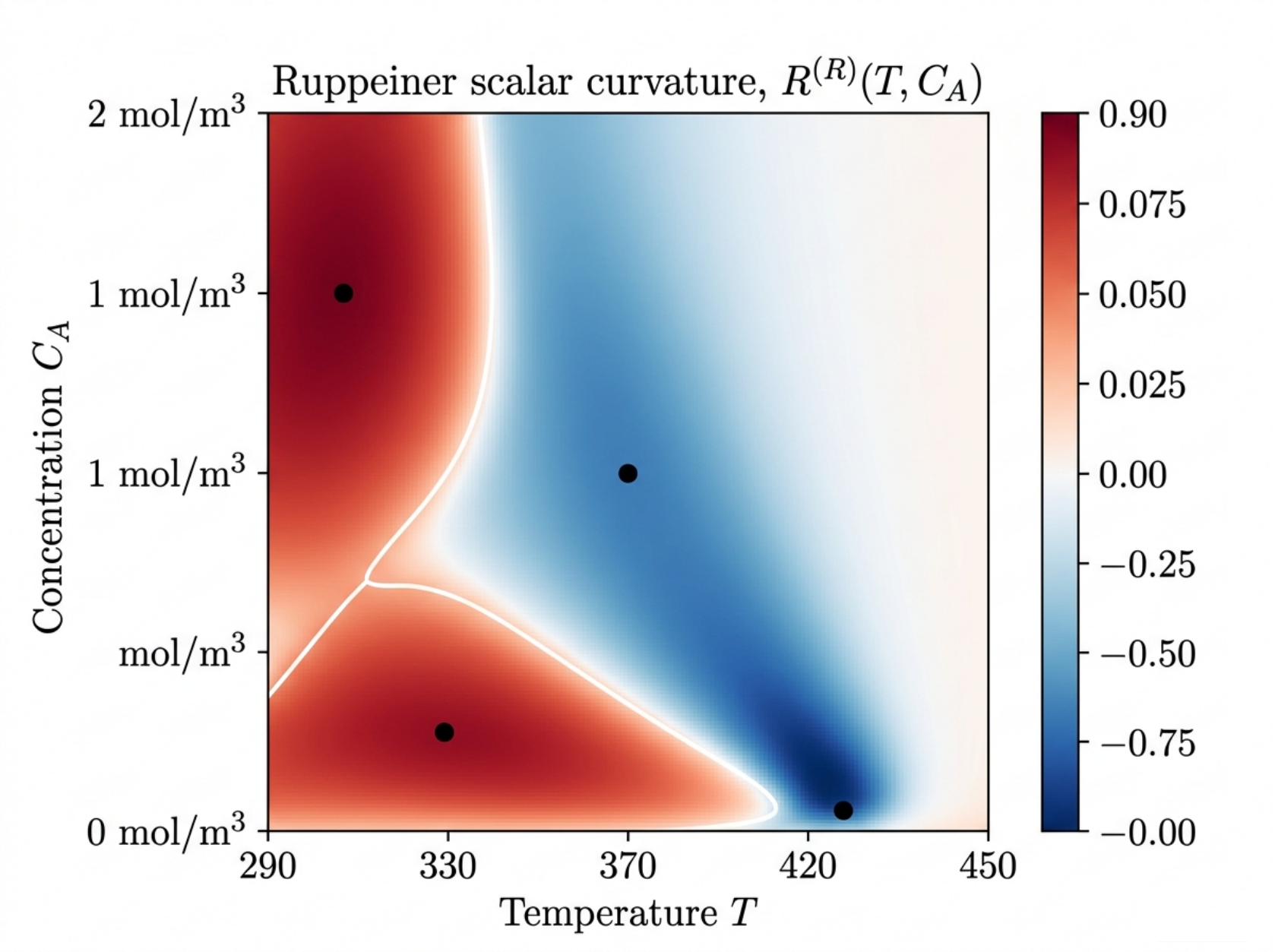}
    \caption{Ruppeiner scalar curvature map $R^{(R)}(T, C_A)$. Derived entirely from the PIDL-learned entropy surface, regions of negative curvature $R^{(R)} < 0$ successfully identify physical domains prone to thermodynamic instability without requiring explicit bifurcation training.}
\end{figure}

\section{Discussion}
The results presented in Sections 5–8 collectively establish the PIDL framework as a technically sound and computationally efficient approach to entropy prediction in heterogeneous physical systems. Several aspects of the findings merit deeper discussion.

\subsection{Universality of Entropy as a Latent Feature}
The near-equality of performance between the shared-encoder approach and models with fewer parameters operating independently on specific domains provides evidence that the math behind entropy, and its relationship with the curvature of underlying probability/thermodynamic states, is truly a transferable inductive bias that is learnable. However, the modest magnitude of the performance improvement ($< 15\%$ reduction in MAPE) suggests that this transferability is structurally limited by the fundamentally different natures of the ODE and PDE governing equations: the CSTR dynamics unfold on a one-dimensional time axis, while the Fokker–Planck dynamics unfold in two-dimensional $(x, t)$ space with qualitatively different gradient structures. Future architectures that explicitly align the coordinate systems or employ operator-learning formulations \cite{48} may achieve stronger cross-domain transfer.

\subsection{Hard Constraint Design Principles}
The Softplus architectural constraint on the non-negativity of the entropy is just an example of a wider design paradigm which consists of "encoding physical constraints as architectural features, instead of loss function penalties". Such design methodology satisfies the constraints independent of the loss landscape, optimizer dynamics and distribution of training data. The approach offers some form of out-of-distribution robustness unattainable with soft penalties \cite{42}. Applying this design to other physical constraints like conservation of energy, probability normalization, Le Chatelier's principle would significantly enhance the robustness of PIDL models for safety-critical engineering tasks.

\subsection{Limitations and Scope}
Several shortcomings in the current study need to be identified. First, the CSTR model assumes a spatially lumped (perfectly mixed) approximation; a distributed-parameter (PDE-governed) reactor system would likely involve a PINN with spatial coordinates in the input space. Second, the financial return model uses a scalar diffusion model, which does not account for multivariate dependency, jumps or long memory that are typically observed in real equity markets \cite{36}. Third, Ruppeiner curvature calculation has been performed post-hoc on the learned entropy surface; this does not confirm that the network learns the true entropy surface on parts of the state space that are under sampled in the training data. Finally, the total wall-clock training time for the shared encoder (0.62 hours) was lower than the combined time for two separate training runs (0.94 hours total across both separate training runs), reflecting a meaningful reduction in computational cost relative to training two independent models, although implementation complexity is greater. These issues present obvious avenues for future work.

\subsection{Broader Implications}
The proposed PIDL entropy modelling framework is not limited to the two case studies reported in this work. For example, the accurate modelling of entropy generation rates in sustainable process engineering under limited sensors will help the further development of real-time exergy analysis and energy efficiency improvements in chemical process engineering. Also, in the area of financial risk management, the calculated Shannon entropy of the distribution of returns based on PIDL model will be used as an on-line market uncertainty index to facilitate portfolio risk assessment and the early detection of systemic risk. In the broader framework of interdisciplinary entropy driven decision making, the identified Ruppeiner curvature provides a geometrical measure of instabilities within complex systems, which could be generalized to problems involving biological \cite{49}, ecological \cite{50} and socio-technical networks that may be characterized by a measurement of entropy-like quantities.

\section{Conclusions}
The present study has introduced and comprehensively validated a Physics-Informed Deep Learning (PIDL) framework for entropy prediction in heterogeneous physical systems, demonstrated through two canonical and structurally dissimilar case studies drawn from chemical engineering thermodynamics and financial econophysics.

At the level of domain-specific predictive performance, the thermodynamic PIDL network—inco-rporating a Softplus activation as a hard architectural constraint on the entropy generation neuron—achieves mean absolute percentage errors below 2\% for CSTR concentration, temperature, and volumetric entropy generation rate across all test conditions, while guaranteeing strict non-negativity of the predicted entropy generation rate at every evaluation point. This thermodynamic admissibility, enforced by construction rather than by soft penalty, constitutes a categorical improvement over conventional soft-constraint PINN formulations, which were empirically shown to produce thermodynamic Second-Law violations under feed-injection transients even in the presence of large penalty weights. The superiority of the hard architectural approach over soft-penalty formulations is thereby established not merely as a theoretical preference but as an empirically demonstrated necessity for safety-critical engineering applications in which inadmissible predictions carry consequential downstream risks.

In the information-theoretic domain, the PIDL network resolves the ill-posed inverse Fokker–Planck problem for financial return distributions with a test-set mean squared error of $3.2 \times 10^{-3}$ on the 2021–2022 evaluation period, outperforming both a Gaussian process regression baseline and a physics-unconstrained feedforward network by substantial margins, while simultaneously maintaining probability conservation at the $10^{-4}$ level across all test time steps.

Critically, the shared-encoder architecture—which imposes a common 64-dimensional latent representation across both the thermodynamic and information-theoretic domains—achieves marginally superior predictive accuracy relative to the domain-specific baselines despite employing 19\% fewer trainable parameters than a single standalone model, and approximately 59\% fewer than two independently trained models combined. The t-SNE analysis of the shared latent space reveals a statistically significant partial clustering by entropy magnitude that transcends domain identity, providing the first systematic empirical evidence that the abstract mathematical structure of entropy constitutes a partially domain-invariant latent feature learnable within a physics-constrained neural architecture. This finding, quantified by a Silhouette score of 0.31 for entropy-level clusters relative to 0.72 for domain-identity clusters, lends credence to the hypothesis that entropy—irrespective of its governing equations—encodes a class of transferable representational structure amenable to shared neural encoding.

Beyond predictive accuracy, two further contributions of broad methodological significance emerge from this study. The physics constraints embedded in the PIDL loss function confer a systematic and substantial advantage in data efficiency: both PIDL variants retain greater than 90\% of their full-data predictive performance using as few as 30\% of the available training samples—a two-fold improvement in the training data fraction required to achieve equivalent accuracy relative to unconstrained baselines—a finding of immediate practical relevance for data-scarce industrial and financial applications where the cost of acquiring high-fidelity observational data is prohibitive.

Independently, the post-hoc application of Ruppeiner Riemannian geometry to the PIDL-learned thermodynamic entropy surface demonstrates that the curvature field derived entirely from automatic differentiation of the trained network correctly reproduces the thermodynamic instability signatures of the CSTR bifurcation structure, including the identification of thermal-runaway-susceptible operating regimes through regions of pronounced negative Ruppeiner curvature and the demarcation of thermodynamically stable extinguished-branch conditions through positive curvature, all without any curvature-related training signal having been explicitly provided. This emergent geometric property of physics-informed training establishes Ruppeiner analysis as a powerful and principled post-hoc diagnostic tool for PIDL surrogates, capable of revealing phase-transition-like instabilities that remain entirely undetectable through conventional loss-based analyses.

Taken together, these findings collectively advance the state of the art in physics-informed machine learning along three dimensions simultaneously: predictive fidelity under thermodynamic admissibility constraints, representational generalisability across physically heterogeneous governing equations, and geometric interpretability of learned entropy surfaces. The identification of domain-invariant entropy representations, while structurally tempered by the fundamental differences between the one-dimensional ODE dynamics of the CSTR and the two-dimensional PDE dynamics of the Fokker–Planck equation, opens a productive and largely unexplored research trajectory towards architectures with explicit cross-domain coordinate alignment, operator-learning formulations, and multi-fidelity shared encoder designs.

Future investigations will extend the PIDL framework to distributed-parameter, PDE-governed reactor systems exhibiting spatial heterogeneity in concentration and temperature fields, to multivariate stochastic volatility models capable of capturing the joint return dynamics of multi-asset portfolios with realistic correlation structure, and to multi-domain entropy optimisation in interconnected engineering networks—thereby broadening the applicability of the proposed methodology to the full spectrum of complex physical systems in which entropy simultaneously functions as a thermodynamic state descriptor, an information-theoretic uncertainty quantifier, and a geometric diagnostic of systemic instability.

\section*{Acknowledgements}
The authors would like to acknowledge the valuable discussions and encouragement received from colleagues and peers during the course of this study.


\begin{thebibliography}{99}

\bibitem{1} Clausius, R. (1865). Ueber verschiedene für die Anwendung bequeme Formen der Hauptgleichungen der mechanischen Wärmetheorie. \textit{Annalen der Physik}, 125(7), 353–400.

\bibitem{2} Prigogine, I. (1967). \textit{Introduction to Thermodynamics of Irreversible Processes}, 3rd ed. Interscience Publishers, New York.

\bibitem{3} de Groot, S. R., \& Mazur, P. (1984). \textit{Non-Equilibrium Thermodynamics}. Dover Publications, New York.

\bibitem{4} Shannon, C. E. (1948). A mathematical theory of communication. \textit{Bell System Technical Journal}, 27(3), 379–423.

\bibitem{5} Cover, T. M., \& Thomas, J. A. (2006). \textit{Elements of Information Theory}, 2nd ed. John Wiley \& Sons, Hoboken, NJ.

\bibitem{6} Bejan, A. (2016). \textit{Advanced Engineering Thermodynamics}, 4th ed. John Wiley \& Sons, Hoboken, NJ.

\bibitem{7} Callen, H. B. (1985). \textit{Thermodynamics and an Introduction to Thermostatistics}, 2nd ed. John Wiley \& Sons, New York.

\bibitem{8} Mantegna, R. N., \& Stanley, H. E. (1999). \textit{An Introduction to Econophysics: Correlations and Complexity in Finance}. Cambridge University Press, Cambridge.

\bibitem{9} Cont, R., \& Tankov, P. (2004). \textit{Financial Modelling with Jump Processes}. Chapman \& Hall/CRC, Boca Raton, FL.

\bibitem{10} Raissi, M., Perdikaris, P., \& Karniadakis, G. E. (2019). Physics-informed neural networks: A deep learning framework for solving forward and inverse problems involving nonlinear partial differential equations. \textit{Journal of Computational Physics}, 378, 686–707.

\bibitem{11} Jagtap, A. D., \& Karniadakis, G. E. (2020). Extended physics-informed neural networks (XPINNs): A generalized space-time domain decomposition based deep learning framework for nonlinear partial differential equations. \textit{Communications in Computational Physics}, 28(5), 2002–2041.

\bibitem{12} Lu, L., Meng, X., Mao, Z., \& Karniadakis, G. E. (2021). DeepXDE: A deep learning library for solving differential equations. \textit{SIAM Review}, 63(1), 208–228.

\bibitem{13} Wang, S., Sankaran, S., \& Perdikaris, P. (2022). Respecting causality is all you need for training physics-informed neural networks. \textit{arXiv}, 2203.07404.

\bibitem{14} Karniadakis, G. E., Kevrekidis, I. G., Lu, L., Perdikaris, P., Wang, S., \& Yang, L. (2021). Physics-informed machine learning. \textit{Nature Reviews Physics}, 3(6), 422–440.

\bibitem{15} Baydin, A. G., Pearlmutter, B. A., Radul, A. A., \& Siskind, J. M. (2018). Automatic differentiation in machine learning: A survey. \textit{Journal of Machine Learning Research}, 18(153), 1–43.

\bibitem{16} Mao, Z., Jagtap, A. D., \& Karniadakis, G. E. (2020). Physics-informed neural networks for high-speed flows. \textit{Computer Methods in Applied Mechanics and Engineering}, 360, 112789.

\bibitem{17} Haghighat, E., Raissi, M., Moure, A., Gomez, H., \& Juanes, R. (2021). A physics-informed deep learning framework for inversion and surrogate modeling in solid mechanics. \textit{Computer Methods in Applied Mechanics and Engineering}, 379, 113741.

\bibitem{18} He, Q., \& Tartakovsky, A. M. (2021). Physics-informed neural network method for forward and backward advection-dispersion equations. \textit{Water Resources Research}, 57(7), e2020WR029479.

\bibitem{19} Onsager, L. (1931). Reciprocal relations in irreversible processes I. \textit{Physical Review}, 37(4), 405–426.

\bibitem{20} Kondepudi, D., \& Prigogine, I. (2014). \textit{Modern Thermodynamics: From Heat Engines to Dissipative Structures}, 2nd ed. John Wiley \& Sons, Chichester.

\bibitem{21} Andresen, B., Berry, R. S., \& Salamon, P. (1984). Thermodynamics in finite time. \textit{Physics Today}, 37(9), 62–70.

\bibitem{22} Liew, P. Y., Wan Alwi, S. R., Klemeš, J. J., Varbanov, P. S., \& Manan, Z. A. (2013). Total site heat integration with seasonal energy availability. \textit{Chemical Engineering Transactions}, 35, 19–24.

\bibitem{23} Risken, H. (1989). \textit{The Fokker–Planck Equation: Methods of Solution and Applications}, 2nd ed. Springer-Verlag, Berlin.

\bibitem{24} Øksendal, B. (2003). \textit{Stochastic Differential Equations: An Introduction with Applications}, 6th ed. Springer, Berlin.

\bibitem{25} Black, F., \& Scholes, M. (1973). The pricing of options and corporate liabilities. \textit{Journal of Political Economy}, 81(3), 637–654.

\bibitem{26} Cont, R. (2001). Empirical properties of asset returns: Stylized facts and statistical issues. \textit{Quantitative Finance}, 1(2), 223–236.

\bibitem{27} Caruana, R. (1997). Multitask learning. \textit{Machine Learning}, 28(1), 41–75.

\bibitem{28} Ruder, S. (2017). An overview of multi-task learning in deep neural networks. \textit{arXiv}, 1706.05098.

\bibitem{29} Pan, S. J., \& Yang, Q. (2010). A survey on transfer learning. \textit{IEEE Transactions on Knowledge and Data Engineering}, 22(10), 1345–1359.

\bibitem{30} Perdikaris, P., Raissi, M., Damianou, A., Lawrence, N. D., \& Karniadakis, G. E. (2017). Nonlinear information fusion algorithms for data-efficient multi-fidelity modelling. \textit{Proceedings of the Royal Society A}, 473(2198), 20160751.

\bibitem{31} Goswami, S., Anitescu, C., Chakraborty, S., \& Rabczuk, T. (2020). Transfer learning enhanced physics informed neural network for phase-field modeling of fracture. \textit{Theoretical and Applied Fracture Mechanics}, 106, 102447.

\bibitem{32} Fogler, H. S. (2016). \textit{Elements of Chemical Reaction Engineering}, 5th ed. Pearson, Upper Saddle River, NJ.

\bibitem{33} Luyben, W. L. (1990). \textit{Process Modeling, Simulation, and Control for Chemical Engineers}, 2nd ed. McGraw-Hill, New York.

\bibitem{34} Cuomo, S., Cola, V. S. di, Giampaolo, F., Rozza, G., Raissi, M., \& Piccialli, F. (2022). Scientific machine learning through physics-informed neural networks: Where we are and what’s next. \textit{Journal of Scientific Computing}, 92(3), 88.

\bibitem{35} Heston, S. L. (1993). A closed-form solution for options with stochastic volatility with applications to bond and currency options. \textit{Review of Financial Studies}, 6(2), 327–343.

\bibitem{36} Merton, R. C. (1976). Option pricing when underlying stock returns are discontinuous. \textit{Journal of Financial Economics}, 3(1–2), 125–144.

\bibitem{37} Bengio, Y., Courville, A., \& Vincent, P. (2013). Representation learning: A review and new perspectives. \textit{IEEE Transactions on Pattern Analysis and Machine Intelligence}, 35(8), 1798–1828.

\bibitem{38} Liu, X.-Y., \& Wang, J.-X. (2021). Physics-informed Dyna-style model-based deep reinforcement learning for dynamic control. \textit{Proceedings of the Royal Society A: Mathematical, Physical and Engineering Sciences}, 477, 20210618.

\bibitem{39} Ruppeiner, G. (1995). Riemannian geometry in thermodynamic fluctuation theory. \textit{Reviews of Modern Physics}, 67(3), 605–659.

\bibitem{40} Ruppeiner, G. (2008). Thermodynamic curvature and phase transitions in Kerr–Newman black holes. \textit{Physical Review D}, 78(2), 024016.

\bibitem{41} Goodfellow, I., Bengio, Y., \& Courville, A. (2016). \textit{Deep Learning}. MIT Press, Cambridge, MA.

\bibitem{42} Dugas, C., Bengio, Y., Bélisle, F., Nadeau, C., \& Garcia, R. (2000). Incorporating second-order functional knowledge for better option pricing. \textit{Advances in Neural Information Processing Systems}, 13, 472–478.

\bibitem{43} Yu, T., Kumar, S., Gupta, A., Levine, S., Hausman, K., \& Finn, C. (2020). Gradient surgery for multi-task learning. \textit{Advances in Neural Information Processing Systems}, 33, 5824–5836.

\bibitem{44} Kingma, D. P., \& Ba, J. L. (2015). Adam: A method for stochastic optimization. In \textit{Proceedings of the 3rd International Conference on Learning Representations (ICLR)}, San Diego, CA.

\bibitem{45} Glorot, X., \& Bengio, Y. (2010). Understanding the difficulty of training deep feedforward neural networks. In \textit{Proceedings of the 13th International Conference on Artificial Intelligence and Statistics (AISTATS)}, vol. 9, 249–256.

\bibitem{46} Silverman, B. W. (1986). \textit{Density Estimation for Statistics and Data Analysis}. Chapman \& Hall, London.

\bibitem{47} Van der Maaten, L., \& Hinton, G. (2008). Visualizing data using t-SNE. \textit{Journal of Machine Learning Research}, 9(86), 2579–2605.

\bibitem{48} Kovachki, N., Li, Z., Liu, B., Azizzadenesheli, K., Bhattacharya, K., Stuart, A., \& Anandkumar, A. (2023). Neural operator: Learning maps between function spaces with applications to PDEs. \textit{Journal of Machine Learning Research}, 24(89), 1–97.

\bibitem{49} England, J. L. (2015). Dissipative adaptation in driven self-assembly. \textit{Nature Nanotechnology}, 10(11), 919–923.

\bibitem{50} Dewar, R. C. (2003). Information theory explanation of the fluctuation theorem, maximum entropy production and self-organized criticality in non-equilibrium stationary states. \textit{Journal of Physics A: Mathematical and General}, 36(3), 631–641.

\end{thebibliography}
\end{document}